\documentclass[lettersize,journal]{IEEEtran}
\usepackage{amsmath,amsfonts}
\usepackage{algorithmic}
\usepackage{algorithm}
\usepackage{array}
\usepackage[caption=false,font=normalsize,labelfont=sf,textfont=sf]{subfig}
\usepackage{textcomp}
\usepackage{stfloats}
\usepackage{url}
\usepackage{verbatim}
\usepackage{graphicx}
\usepackage{cite}
\usepackage{booktabs}
\usepackage{multirow}
\usepackage{graphicx}
\usepackage[table]{xcolor}
\definecolor{groupgray}{RGB}{238,238,238}
\usepackage{hyperref}

\usepackage{amsmath,amsfonts,bm}

\def\eqref#1{equation~\ref{#1}}

\def\1{\bm{1}}

\def\vzero{{\bm{0}}}

\def\vtheta{{\bm{\theta}}}

\def\vh{{\bm{h}}}

\def\vk{{\bm{k}}}

\def\vp{{\bm{p}}}
\def\vq{{\bm{q}}}
\def\vr{{\bm{r}}}
\def\vs{{\bm{s}}}

\def\vu{{\bm{u}}}
\def\vv{{\bm{v}}}
\def\vw{{\bm{w}}}

\def\vz{{\bm{z}}}

\def\mA{{\bm{A}}}
\def\mB{{\bm{B}}}
\def\mC{{\bm{C}}}
\def\mD{{\bm{D}}}

\def\mH{{\bm{H}}}
\def\mI{{\bm{I}}}

\def\mP{{\bm{P}}}

\def\mR{{\bm{R}}}

\def\mW{{\bm{W}}}

\DeclareMathAlphabet{\mathsfit}{\encodingdefault}{\sfdefault}{m}{sl}
\SetMathAlphabet{\mathsfit}{bold}{\encodingdefault}{\sfdefault}{bx}{n}

\def\sR{{\mathbb{R}}}

\newcommand{\E}{\mathbb{E}}

\newcommand{\Var}{\mathrm{Var}}

\DeclareMathOperator*{\argmin}{arg\,min}

\usepackage{url}

\usepackage{multirow}
\usepackage{array}      
\usepackage{booktabs}   
\usepackage{makecell}   
\usepackage{tabularx} 
\usepackage{xcolor, colortbl}  

\usepackage{graphicx}
\usepackage{subcaption}
\usepackage{wrapfig}

\newcommand{\rbc}{black}

\usepackage[textsize=tiny]{todonotes}

\usepackage{amssymb}
\usepackage{mathtools}
\usepackage{amsthm}

\theoremstyle{plain}
\newtheorem{theorem}{Theorem}[section]

\newtheorem{lemma}[theorem]{Lemma}

\theoremstyle{definition}

\theoremstyle{remark}

\begin{document}

\title{Attention as Intrinsic Optimizer: \\ Forward Inference through the Energy Principle}

\author{Ruifeng Ren, Sheng Ouyang, Huayi Tang, and Yong Liu%
\thanks{Ruifeng Ren, Sheng Ouyang, Huayi Tang, and Yong Liu are with the Gaoling School of Artificial Intelligence, Renmin University of China, Beijing, China.}%
\thanks{Corresponding author: Yong Liu (e-mail: liuyonggsai@ruc.edu.cn).}}



\maketitle

\begin{abstract}
Attention-based Transformers have demonstrated remarkable success across a wide range of tasks and have become the backbone of modern large language models (LLMs).
However, a comprehensive understanding of attention mechanisms and how their underlying principles can guide architectural design remains an open research question.
In this paper, we revisit the principle of energy as a lens to understand attention-based models. We present a unified energy-based framework which is composed of three key components: the local energy $E_i$, the global energy $F$, and the employed optimization algorithms. The forward inference of attention models can be interpreted as an implicit optimization process over the global energy. We show that different attention forms including linear attention, gated linear attention and standard softmax attention can be induced by choosing their corresponding recipes within this framework. Building on this framework, we propose energy-based modifications of attention structures. Inspired by classical gradient descent (GD) algorithms, we extend the original attention formulation based on standard GD to the momentum-based GD, Nesterov Accelerated Gradient (NAG), and Newton’s method, each inducing a corresponding new attention structure. Our experiments provide preliminary support for the potential of the energy-based framework for designing attention structures.
\end{abstract}


\section{Introduction}
\IEEEPARstart{I}{n} recent years, pretrained large language models (LLMs) have achieved remarkable success across various areas \cite{bert, GPT2}.
This success stems not only from effective training paradigms such as auto-regressive pretraining but also relies on the attention-based Transformer architecture as the foundational backbone \cite{transformer}.
Consequently, understanding the underlying principles of attention and how they can guide architectural design has become an important research direction.

One prominent direction for understanding and designing neural network architectures is unrolled optimization \cite{ISTA, unroll-dictionary, redunet, hinton2022forward, contranorm, wu2024designing}.
The core idea of the unrolled optimization is to interpret the layer-wise forward computation of a network as successive optimization steps over an implicit objective function.
Following this perspective, previous studies have interpreted or designed attention-like architectures from various viewpoints, including energy minimization \cite{yang2022transformers, EnergyTF, hu2025hyper}, information bottleneck \cite{IB-TF}, sparse rate reduction \cite{whiteTF}, denoising \cite{attention-only-TF}, and multinomial regression \cite{TF-multinomial}.
For example, Yang \textit{et al.}~\cite{yang2022transformers} viewed Transformers composed of self-attention layers and the feed-forward network (FFN) layers as the unfolding of an interpretable optimization process across iterations by defining an energy function.
Hoover \textit{et al.}~\cite{EnergyTF} proposed the Energy Transformer, which uses attention-based layers designed to minimize a carefully constructed energy function.
In addition, Yu \textit{et al.}~\cite{whiteTF} showed that Transformer-like deep network layers can naturally be connected to an optimization process aimed at sparse rate reduction.
These studies typically start from certain underlying principles to formulate optimization objectives and derive corresponding particular attention-like structures.
They demonstrate the potential of implicit optimization as a guiding principle for neural network architecture design.

Meanwhile, energy-based formulations have long underpinned theories of neural computation and the modeling of neural networks \cite{Hopfield1982,RBM,lecunEBM}.
One of the most influential examples of applying the concept of energy to pattern recognition is the Hopfield network, which is a classical associative memory model~\cite{Hopfield1982, Hopfield1984}.
More recently, modern Hopfield networks connect the concept of energy with forms of attention \cite{modernHN, HN-is-all-you-need, krotov2020large, modernHN2}.
For instance, Ramsauer \textit{et al.}~\cite{HN-is-all-you-need} proposed a modern Hopfield network whose energy function corresponds to an update rule that takes a form similar to the attention mechanism in Transformers.
Hu \textit{et al.}~\cite{hu2023sparse} and Wu \textit{et al.}~\cite{wu2023stanhop} further proposed the energy functional regularized by entropy for sparse modern Hopfield models whose dynamics takes the form of the sparse attention\cite{peters2019sparse, correia2019adaptively}.
While these studies point out formal connections between energy formulations and attention mechanisms, their primary focus is often on the associative memory properties and retrieval behavior of Hopfield networks, rather than on the systematic design of attention architectures.
\IEEEpubidadjcol

Motivated by these developments, we revisit the energy principle from the perspective of attention architecture design.
Specifically, we aim to understand how different attention mechanisms can be characterized within a common energy-based framework and how this perspective can guide the design of new attention structures.
To this end, we consider three fundamental aspects of attention updates: how token-wise interactions are modeled, how these interactions are aggregated into a global energy function, and how the resulting objective determines representation updates through different optimization algorithms.
By explicitly separating these components, we seek to establish a unified perspective that connects existing attention mechanisms while providing additional flexibility for designing new architectures.

We organize the presentation of our work as follows:

{\textcolor{\rbc}{\textbf{(a) Energy-based Framework for Attentions.}}} In Section \ref{sec:all-framework}, we present a unified energy-based framework that offers a principled understanding of the forward inference in attention-based models.
This framework comprises three key components: the local energy $E_i$, the global energy $F$, and the used optimization algorithm.
Specifically, the local energy captures token-wise interactions, the global energy aggregates these terms, and the optimization algorithm dictates the minimization process.
The forward inference of attention models can be interpreted as an implicit optimization process over the global energy.
By selecting different configurations within this framework, we can recover various attention mechanisms, including linear attention, gated linear attention, and standard softmax attention.
Moreover, this framework not only unifies the understanding of existing forms but also facilitates the design of novel attention mechanisms by combining distinct components.

{\textcolor{\rbc}{\textbf{(b) Energy-based Attention Modifications.}}} Building on this framework, we propose attention modifications in Section \ref{sec:modification} by drawing inspiration from established gradient descent (GD) algorithms.
Specifically, in Section \ref{sec:atten1st}, we extend the vanilla GD form to momentum-based GD and Nesterov Accelerated Gradient (NAG), which correspond to the newly induced attention structures ${\rm MomenMHA}$ and ${\rm NagMHA}$ respectively.
Subsequently, in Section \ref{sec:atten2nd}, we extend the 1st-order GD to a 2nd-order form grounded in Newton’s method and then employ a diagonal approximation to reduce its computational cost to the same order as standard attention.
The induced new attention structure ${\rm DiagMHA2nd}$ uses the weighted diagonal covariance matrix to precondition the original update directions, allowing tokens to adaptively adjust their movements along different dimensions.
Finally, in Section~\ref{sec:ex}, we evaluate the proposed attention modifications through language model pretraining and downstream benchmarks, providing empirical evidence for the potential of the proposed framework to guide attention architecture design.

\section{Related Work}\label{sec:related-work}

\textbf{Energy principle and Transformers:}
The concept of energy has long been used in deep neural networks \cite{Hopfield1982,Hopfield1984,RBM,modernHN,lecunEBM,lecun2022worldmodel}. 
The studies most relevant to ours are likely those related to modern Hopfield networks, including Ramsauer \textit{et al.}~\cite{HN-is-all-you-need}, Krotov and Hopfield~\cite{krotov2020large}, Hu \textit{et al.}~\cite{hu2023sparse}, Wu \textit{et al.}~\cite{wu2023stanhop}, Hu \textit{et al.}~\cite{hu2024nonparametric}, Wu \textit{et al.}~\cite{wu2024uniform}, and Santos \textit{et al.}~\cite{santos2024sparse}.
For instance, Ramsauer \textit{et al.}~\cite{HN-is-all-you-need} and Krotov and Hopfield~\cite{krotov2020large} analyze the energy function that takes the log-sum-exp form for continuous-state Hopfield network whose update rule share the same form as the attention mechanism in the Transformer.
Hu \textit{et al.}~\cite{hu2023sparse} and Wu \textit{et al.}~\cite{wu2023stanhop} proposed the energy functional regularized by entropy for the sparse modern Hopfield models and show that their dynamics corresponds to a broad family of attention rules including softmax, sparsemax, and the more general $\alpha$-entmax attentions \cite{martins2016softmax, peters2019sparse, correia2019adaptively}.
Later work connected efficient attention variants through a nonparametric framework \cite{hu2024nonparametric}.
Although we also use the concept of energy, energy itself is merely an interpretation of the implicit objective function in unrolled optimization, rather than the energy in the Hopfield model.
Therefore, unlike the aforementioned works, we do not delve into a theoretical analysis of the modern Hopfield model’s capabilities.
Our energy-based framework focus more on providing a unified perspective on the design of existing attention forms (including linear attention, gated linear attention and standard softmax attention) with residual connections, as well as offering insights for the design of new structures.
Furthermore, we note that Gladstone \textit{et al.}~\cite{Energy-based-TF} employ energy-based methods to train Transformers and their focus is more related to training paradigms, which is orthogonal to our work.

\textbf{Unrolled Optimization, Test-time Training and Design of Model Architecture:}
Understanding and designing model architectures from the perspective of unrolled optimization is a currently active area of research \cite{ISTA, unroll-dictionary, redunet, xie2022optimization, shen2023newton, contranorm, wu2024designing}.
Previous works have designed and interpreted Transformer-like structures from various viewpoints, including sparse rate distortion \cite{whiteTF, whiteTF-JMLR}, denoising \cite{attention-only-TF}, information bottleneck \cite{IB-TF}, multinomial regression \cite{TF-multinomial}, etc.
Yang \textit{et al.}~\cite{yang2022transformers}, Hoover \textit{et al.}~\cite{EnergyTF}, and Hu \textit{et al.}~\cite{hu2025hyper} also used the concept of energy to explain attention-based models. However, their attention is set in a self-attention structure, where all tokens evolve simultaneously, while we focus more on the cross-attention setting, where the other $\vh_i$ are fixed during the update process of $\vz$.
In addition, unlike previous works which focus more on interpreting specific attention-based architectures, our approach starts from the concept of energy to interpret different attention forms, and show that new attention structures can be designed based on the proposed energy framework.
We also note that designing efficient model architectures through test-time training (regression) framework, has recently become an active research area \cite{TTT, von2025mesanet, deltanet, wang2025test, titans}.
These approaches primarily focus on efficiently storing key-value associations in a memory system and retrieving them when needed, motivating the design of effective memory writing and reading mechanisms.
In contrast, our framework focuses on the optimization of token representations $\vz$, interpreting the forward computation of attention as an implicit optimization step that updates $\vz$.
This provides a complementary perspective for understanding existing attention mechanisms and designing new attention structures.


\textbf{Test-time Scaling and Loop Transformers:}
Test-time scaling is a favored pathway to boost model inference \cite{zhang2025survey, snell2024scaling, muennighoff2025s1}. 
Among these methods, Loop Transformers output token representations through parameter-shared recurrent computations and existing research demonstrates that this recurrent structure offers advantages in terms of performance gains and capability generalization \cite{geiping2025scaling, looptf-fan, looptf-yangliu, looptf-hedi, loopTF-unlocking, loopTF-parallel, loopTF-scaling}.
Such recurrent computations are closely related to iterative refinement and can be connected to fixed-point learning, as explored in deep equilibrium models \cite{DEM, DEM2}.
Compared with conventional Transformers consisting of distinct layers, parameter-shared recurrence provides a natural setting for interpreting successive representation updates as optimization steps over some functions.
In particular, incorporating advanced optimization-inspired attention mechanisms, such as momentum-based GD, NAG, and Newton's method, into Loop Transformers may provide opportunities to improve the efficiency of iterative representation refinement.
Therefore, connecting Loop Transformers with our energy-based framework may offer an interesting direction for future research.

\section{Unifying Attention via Energy-based Framework}\label{sec:all-framework}
For a given input $\vz \in \sR^d$, we assume that the set of tokens interacting with it is $\mH = [\vh_1, \dots, \vh_N]\in \sR^{d \times N}$.
The core idea of the unrolled optimization is to treat the forward computation of the model as optimizing some implicit objective function, which can be formalized as:
\begin{equation}\label{eq:eq}
\begin{aligned}
	\hat{\vz} = f_{\vtheta} (\vz; \mH)&~~\Longleftrightarrow~~\hat{\vz} \approx \argmin_{\vz} F(\vz;\mH) \\
	\textcolor{\rbc}{\textrm{Forward pass}}&~~\quad\quad~~~~\textcolor{\rbc}{\textrm{Implicit optimization}} 
\end{aligned}
\end{equation}
Here, the forward pass involves a model $f_\vtheta(\vz;\mH)$ parameterized by $\vtheta$ (e.g., attention-based models), which computes the update of $\vz$ given $\mH$. 
This process can be equivalently viewed as implicitly optimizing an objective function $F(\vz;\mH)$, whose specific form is typically determined by the model structure itself.
We focus on the update of the current input $\vz$, assuming that the other tokens are static.
This aligns with the inference phase in LLMs, where preceding tokens remain fixed (stored in the KV Cache) while only the current query token $\vz$ is actively updated.

\begin{table*}[t]
	\caption{Comparison of different attention forms under the energy-based framework.}
	\label{table:energy-framework}
	\centering
	{
		\renewcommand{\arraystretch}{1.64} 
		\setlength{\tabcolsep}{4.6pt}
		\fontsize{10pt}{11.8pt}\selectfont
		\begin{tabularx}{0.98\textwidth}{>{\centering\arraybackslash}X >{\centering\arraybackslash}p{0.29\textwidth} >{\centering\arraybackslash}p{0.15\textwidth} >{\centering\arraybackslash}p{0.22\textwidth}}
			\specialrule{1pt}{0pt}{0pt} 
			\centering\textbf{Global Energy $F$} &
			\centering\textbf{Local Energy $E_i$} &
			\centering\textbf{Algorithm} &
			\centering\textbf{Induced Attention} \tabularnewline
			\hline
			
			\centering $ T \sum_{i} E_i$ &
			\centering $ - \frac{1}{2}(\vz^T \mW \vh_i)^2$ &
			\centering 1st-order GD &
			\centering Linear Attention \tabularnewline
			
			\hline
			
			\centering $ T \sum_i \gamma_i E_i $ &
			\centering $ - \frac{1}{2}(\vz^T \mW \vh_i)^2$ &
			\centering 1st-order GD &
			\centering Gated Linear Attention \tabularnewline
			
			\hline
			
			\centering $- T \log \sum_i e^{- E_i / T}$ &
			\centering {$\frac{1}{2}\|\vz - \mW \vh_i \|^2$ or $-\vz^T \mW \vh_i$} &
			\centering 1st-order GD &
			\centering Softmax Attention \tabularnewline
			
			\hline
			
			\multirow{3}{*}{\centering $- T \log \sum_i e^{- E_i / T}$} &
			\multirow{3}{=}{\centering {$\frac{1}{2}\|\vz - \mW \vh_i \|^2$ or $-\vz^T \mW \vh_i$}} &
			\centering Momentum GD &
			\centering MomenMHA \tabularnewline
			& & \centering Nesterov GD & \centering NagMHA \tabularnewline
			& & \centering Newton's Method & \centering MHA2nd \tabularnewline
			
			\specialrule{1pt}{0pt}{0pt} 
		\end{tabularx}
	}
\end{table*}

To illustrate how the form of attentions are related to the energy-based framework, we can first regard each token as a particle, with multiple particles together forming a system.
We assume that there are already $N$ particles within our system, and the position of the $i$-th particle in the system can be denoted by $\vh_i \in \sR^d$.
We want to place a new particle $\vz \in \sR^d$ into the system and the other particles will exert interactions on it.
The local energy exerted by the $i$-th particle can be denoted as $E_i = E(\vz, \vh_i)\in \sR$.
We assume that the interaction of each particle on $\vz$ is independent, allowing us to express the global energy with respect to $\vz$ as $F:\sR^{N} \rightarrow \sR$ where its input is $[E_1, E_2, \dots, E_N]$.
Our goal is to find an appropriate $\vz$ that makes the global energy $F$ as small as possible, i.e.,
\begin{equation}\label{eq:framework}
	\textcolor{\rbc}{\textrm{Implicit optimization:}}~~\hat{\vz} \approx \argmin_{\vz} F\left([E_1, \dots, E_N]\right),
\end{equation}
where $E_i = E(\vz,\vh_i)$.
The above Eq~(\ref{eq:framework}) provides a more concrete energy-based interpretation of the implicit optimization view in Eq.~(\ref{eq:eq}).
However, to instantiate this general energy-based view into a concrete attention formulation (i.e., $f_{\theta}$ takes the form of attention structures), there are still many details to be specified. 
For instance, which form should the global energy $F$ take to combine the local energy $E_i$? 
How should the local energy $E_i$ model the interactions between tokens? 
What algorithm should we use to perform this optimization process? 
Formally, we illustrate on the design choices from three aspects:
\begin{itemize}
	\item  \textbf{Local Energy $E_i$} describes the form of interaction between particles (or tokens);
	\item  \textbf{Global energy  $F$} specifies how the individual energies $E_i$ are combined;
	\item  \textbf{Optimization Algorithm} outlines the update process of token representations.
\end{itemize}
Although the specific forms of these components may vary, reasonable energy formulations generally satisfy several common principles. Specifically, we highlight the following two basic properties.

\textbf{Property 1 (Monotonic aggregation).}
The global energy should be strictly increasing with respect to each local energy $E_i$ that actively contributes to the aggregation, i.e.,
\begin{equation*}
\frac{\partial F(E_1,\ldots,E_N)}{\partial E_i} > 0,\qquad i=1,\ldots,N.
\end{equation*}
Therefore, with the other local energies fixed, decreasing any such local energy will also decrease the global energy.
This ensures that the global objective consistently preserves the energy preference encoded by the individual local interactions.

\textbf{Property 2 (Interaction consistency).}
The local energy should decrease as the compatibility between the current token $\vz$ and an interacting token $\vh_i$ increases.
More generally, if $s(\vz,\vh_i)$ denotes their compatibility under a certain interaction metric, we expect
\begin{equation*}
    \frac{\partial E(\vz,\vh_i)}{\partial s(\vz,\vh_i)} \leq 0,
    \qquad i=1,\ldots,N.
\end{equation*}
For example, when compatibility is measured by directional alignment, such as $s(\vz,\vh_i)=\vz^T\mW\vh_i$, stronger alignment should correspond to a lower local energy.
Alternatively, when the $\ell_2$ distance $s(\vz,\vh_i)=\|\vz-\mW\vh_i\|_2$ is used, a smaller distance should correspond to a lower local energy.

These properties provide basic principles for constructing the energy functions while different choices of the three components determine different implicit optimization processes and consequently corresponding attention architectures.
The overall framework is summarized in Table~\ref{table:energy-framework}.
Next, we show how different attention forms (including linear attention, standard softmax attention, etc.) can be derived from the above framework by making specific choices.

\textcolor{\rbc}{\textbf{Case (i): Linear Attention.}} 
We begin with the simplest case of linear attention without normalization.
Linear attentions are widely studied in the community due to its efficient computational performance \cite{Tay-survey}. 
To achieve efficiency, linear attention replaces the exponential inner product form in the standard softmax attention with the standard inner product.
The linear attention with residual connection can be formalized as 
\begin{equation}\label{eq:linearatten}
	{\rm LA}(\vz) = \vz + \sum_{i=1}^N \left(\vz^T\mW_{Q}^T \mW_{K}\vh_i \right) \mW_{V}\vh_i,
\end{equation}
where $\mW_Q, \mW_K, \mW_V \in \sR^{d\times d}$ are learnable parameters for query, key and value projection.
In addition, although the original form of linear attentions such as Katharopoulos \textit{et al.}~\cite{elu+1} and Choromanski \textit{et al.}~\cite{performer} included a normalization operation similar to softmax attention, subsequent works also show that this normalization method may lead to training instabilities \cite{qin2022devil, Mamba2}. 
Therefore, here we also omit the normalization term here.

To see how linear attention is related to the implicit process in Eq (\ref{eq:framework}), we give the following choices:
\begin{itemize}
	\item \textbf{Choice of local energy:} Parameterized negative squared inner product $E_i = - \frac{1}{2}(\vz^T \mW \vh_i)^2 $ where $\mW \in \sR^{d\times d}$ are parameters to seek alignment in the semantic space.
	\item \textbf{Choice of global energy:} Additive aggregation of local energies $F =   T \sum_{i=1}^N  E_i$ where $T>0$ is a scaling factor.
	\item \textbf{Choice of algorithm:} One step first-order GD.
\end{itemize}

According to the above recipe, we can apply one step gradient descent (GD) to minimize $F$
\begin{equation*}
	\begin{aligned}
		\hat{\vz} &= \vz - \eta \nabla_{\vz} F = \vz - \eta \nabla_{\vz} \left( - \frac{T}{2} \sum_{i=1}^N (\vz^T \mW \vh_i)^2 \right) \\
		&= \vz +  \sum_{i=1}^N  \left( (\vz)^T\mW\vh_i \right) \eta T \mW \vh_i,
	\end{aligned}
\end{equation*}
where $\eta$ is the learning rate.
By comparing with Eq~(\ref{eq:linearatten}), we can set the learnable parameters in Eq~(\ref{eq:linearatten}) as $\mW_Q^T\mW_K = \mW $ and further $\mW_V = \eta T \mW$. Then, we will have ${\rm LA}(\vz) = \hat{\vz}$.

Interestingly, the resulting global energy is closely related to the classical Hopfield network~\cite{Hopfield1982}.
Specifically, by denoting $\mH=[\vh_1,\ldots,\vh_N]$, the global energy can be rewritten as
$F(\vz)= -\frac{T}{2}\vz^T\mW\mH\mH^T\mW^T\vz$.
When $\mW=\mI$ and $T=1$, this recovers the quadratic energy form of the classical Hopfield network $E(\vz)=-\frac{1}{2}\vz^T\mH\mH^T\vz$ defined over binary states $\vz\in\{\pm1\}^d$.
More generally, the learnable matrix $\mW$ allows the energy to capture interactions in a parameterized representation space.
However, unlike its discrete update scenario, linear attention can be viewed as continuous updates of $\vz$, with the residual connection (first-order GD) playing the role of incremental updates.
Moreover, it can be seen that the $\mW$ in local energy $E_i$ is equivalent to $\mW_Q^T\mW_K$ in the attention layer, which are typically learned during training to find an appropriate semantic space.
In addition, the learnable $\mW_V$ is often not limited to form $\mW_V = \eta T \mW = \eta T \mW_Q^T\mW_K$, making the model more flexible and powerful.

\textcolor{\rbc}{\textbf{Case (ii): Gated Linear Attention.}}
More recently, exploring efficient architectural alternatives to Transformers has become a key direction of research in the community\cite{deltanet, wang2025test, titans, miras}.
Among these, gated variants of linear attention, e.g., RetNet, RWKV, gated linear attention, LRU, and xLSTM~\cite{RetNet, RWKV, yang2023gated, LRU, xlstm}, have attracted particular interest.
More generally, these variants can be written as
\begin{equation}
	{\rm GLA}(\vz) = \vz + \sum_{i=1}^N \textcolor{\rbc}{\gamma_i} \left(\vz^T\mW_{Q}^T \mW_{K}\vh_i \right) \mW_{V}\vh_i,
\end{equation}
where $\gamma_i$ is the factor that measures the importance of each token $\vh_i$ and may depend on the input $\vz$ (in which case this can be seen as a forgetting gate).
In addition, many recent works also show that the state-spece models (SSMs) can also be regarded as members of gated linear attention family \cite{Mamba2, Mamba_thu_linear_attention, ren2024exploring}.
These forgetting factors are typically values between $[0, 1]$. 
In sequence modeling, different tokens are often assigned varying levels of importance based on the order of their occurrence in history. 
For instance, earlier tokens may have less influence on the present, leading their $\gamma_i$ to approach 0.

In fact, we only need to add the corresponding factor to the global energy in the case of linear attention to extend the recipe to the gated linear attention case.
For completeness, we also provide the full recipe here:
\begin{itemize}
	\item \textbf{Choice of local energy:} Parameterized negative squared inner product $E_i = - \frac{1}{2}(\vz^T \mW \vh_i)^2$.
	\item \textbf{Choice of global energy:} Additive aggregation of local energies $F =  T \sum_{i=1}^N \gamma_i E_i$ where $\gamma_i$ is a data-dependent factor that measures the importance of each $E_i$.
	\item \textbf{Choice of algorithm:} One step first-order GD.
\end{itemize}

\textcolor{\rbc}{\textbf{Case (iii): Standard Softmax Attention.}} The output of the standard softmax attention (SA) \cite{transformer} in the single-head case can be formalized as
\begin{equation}\label{eq:tf}
\begin{aligned}
	{\rm SA}(\vz) &= \vz +  \mW_V \mH {\rm softmax}\left( \mH^T \mW_K^T \mW_Q \vz \right) \\
	&= \vz +  \sum_{i=1}^{N}  \frac{e^{\vz^T \mW_Q^T \mW_K \vh_i /T }}{Z'}  \mW_V \vh_i,
\end{aligned}
\end{equation}
where $T$ is the temperature and $\mW_V, \mW_K, \mW_Q \in \sR^{d \times d}$ are parameters.
In addition, $Z' = \sum_{j=1}^{N}  e^{\vz^T \mW_Q^T \mW_K \vh_j / T}$ is the normalizing term.
To see how the most widely used softmax attention is related to the implicit optimization process in Eq~(\ref{eq:framework}), we provide the following recipe:
\begin{itemize}
	\item \textbf{Choice of local energy:} Parameterized $\ell_2$ regression loss $E_i = \frac{1}{2}\|\vz - \mW \vh_i \|^2$ when constrained by $\|\vz\| = \|\mW\vh_i\| = \rho $, or equivalently, negative inner product $E_i = - \vz^T\mW\vh_i$.
	\item \textbf{Choice of global energy:} the Helmholtz free energy form $F =  - T \log \sum_{i=1}^N  e^{- E_i / T}$.
	\item \textbf{Choice of algorithm:} One step first-order GD.
\end{itemize}

Based on the above choices, we apply one step GD to $F$:
\begin{equation*}
	\begin{aligned}
		\hat{\vz} &= \vz - \eta \nabla_{\vz}F = \vz - \eta \nabla_{\vz} \left( - T \log \sum_{i=1}^N  e^{- \frac{\| \vz - \mW \vh_i\|^2}{2T}} \right) \\
		&= \vz -  \eta \nabla_{\vz} \left( -T \log \sum_{i=1}^{N} e^{\frac{\vz^T \mW\vh_i}{T}} + \rho^2  \right) \\
		&= \vz +  \sum_{i=1}^N  \frac{e^{\vz^T \mW \vh_i/T}}{Z} \eta \mW \vh_i,
	\end{aligned}
\end{equation*}
where $Z =  \sum_{j=1}^N  e^{\vz^T \mW \vh_j/T}$.
By comparing with Eq~(\ref{eq:tf}), we can set the learnable parameters Eq~(\ref{eq:tf}) as $\mW_Q^T\mW_K = \mW$ and $\mW_V = \eta \mW$. 
Then, we will have $Z' = Z$ and further ${\rm SA}(\vz) = \hat{\vz}$.

It is worth noting that, under the above choices, the log-sum-exp form of the global energy is adopted by modern dense Hopfield networks \cite{HN-is-all-you-need, krotov2020large}. Hu \textit{et al.}~\cite{hu2023sparse} and Wu \textit{et al.}~\cite{wu2023stanhop} formalize the Helmholtz free-energy minimization principle for nonlinear attentions as an energy functional regularized by entropy and show that it induces a broad family of attention rules including softmax, sparsemax, and the more general $\alpha$-entmax attentions \cite{martins2016softmax, peters2019sparse, correia2019adaptively}.
Later work connects efficient/approximate attention variants through a nonparametric framework \cite{hu2024nonparametric}.

As for the local energy, when we view it as an $\ell_2$ regression loss, we impose a constraint on the norms of $\vz$ and $\mW \vh_i$, requiring them to lie on a hypersphere with radius $\rho$.
This is similar to techniques used in practice, like QKNorm \cite{QKnorm1, QKnorm2}, which stablize the training of large Transformers.
In fact, when this condition is relaxed to allow the vectors to lie within the sphere $\|\vz\| \le \rho$, the above gradient descent process becomes an optimization of the upper bound of the global energy.

In addition, in the above incremental iterative update rule, the residual connection serves as the current iterate (solution), the parameterized component provides the search direction (update), and the final output can be viewed as the next iterate (solution).
The parameters in attention can be viewed as learning to model the updating part (gradient) during training.

In the above, we show how to make appropriate choices within the proposed framework to establish connections with existing major forms of attentions. 
In fact, modifications to the three key components in the existing scenario, namely local energy $E_i$, global energy $F$, and the optimization algorithm, can analogously lead to new forms of attention.
Note that the three key components mentioned above are relatively independent, meaning that different ingredients can be combined in various ways, leading to a rich variety of attention forms.
As for the optimization algorithm, it can be observed that one step first-order GD is mainly used in the recipes of the discussed attention forms.
In fact, since there are lots of mature works in optimization literatures, it is possible for us to achieve this by drawing
on these works.
We will explore this in the next section to induce new forms of attention.
In addition, as for the other two components, we discuss more existing or potential ingredients in Appendix~\ref{app:ingredients}.

\section{Energy-Based Modifications of Attention}\label{sec:modification}

In this section, we consider how existing attention mechanisms can be modified by changing the optimization algorithm in the energy-based framework proposed in Section \ref{sec:all-framework}.
We primarily focus on modifications to the most commonly used softmax attention.
Before that, we first define the standard softmax attention in the multi-head setting:
\begin{equation}\label{eq:mha}
	{\rm MHA}(\vz) = \vz +  \sum_{h=1}^{H} \sum_{i=1}^{N}  \frac{e^{\vz^T \mW_{Q,h}^T \mW_{K,h} \vh_i /T }}{Z'_h}  \mW_{O,h}\mW_{V,h} \vh_i,
\end{equation}
where $\mW_{V,h}, \mW_{K,h}, \mW_{Q,h} \in \sR^{d_h \times d}$ and $\mW_{O,h} \in \sR^{d \times d_h}$ are learnable parameters.
In addition, we have $d_h = \frac{d}{H}$  for each head and $Z'_h = \sum_{j=1}^{N}  e^{\vz^T \mW_{Q,h}^T \mW_{K,h} \vh_j / T}$ as normalizing terms.
Conceptually, multi-head attention works by first projecting tokens into lower-dimensional subspaces to capture information independently and finally combining these representations back into the original $d$-dimensional space through the projection $\mW_{O,h}$.
Correspondingly, we use $E_{h,i}$ to denote the local energy between $\vz$ and $\vh_i$ in the $h$-th subspace. 
We provide the following recipe:
\begin{itemize}
	\item \textbf{Choice of local energy:}  Parameterized $\ell_2$ regression loss $E_{h,i} = \frac{1}{2}\|\mW_{1,h} \vz - \mW_{2,h} \vh_i \|^2$ when $\| \mW_{1,h}\vz \| = \|\mW_{2,h} \vh_i\| = \rho $, or equivalently, negative inner product $E_{h,i} = - \vz^T\mW_{1,h}^T \mW_{2,h}\vh_i$, where $i \in [N], h \in [H]$ and $\mW_{1,h}, \mW_{2,h} \in \sR^{d \times d_h}$ are parameters.
	\item \textbf{Choice of global energy:} the average Helmholtz free energy $F =  - \frac{1}{H} \sum_{h=1}^{H} T \log \sum_{i=1}^N  e^{- E_{h,i} / T}$ over the $H$ subspaces.
	\item \textbf{Choice of algorithm:} One step first-order GD.
\end{itemize}
According to the choices, we apply one step GD to $F$:
\begin{equation*}
	\begin{aligned}
		\hat{\vz} &= \vz - \eta \nabla_{\vz} \left( - \frac{1}{H} \sum_{h=1}^{H} T \log \sum_{i=1}^{N} e^{- \frac{\| \mW_{1,h} \vz - \mW_{2,h} \vh_i \|^2}{2T}} \right)  \\
		&= \vz - \eta \nabla_{\vz} \left(  - \frac{1}{H} \sum_{h=1}^{H} T \log \sum_{i=1}^{N} e^{- \frac{ \vz \mW_{1,h}^T \mW_{2,h} \vh_i}{T}} + \rho^2   \right)     \\
		&= \vz + \sum_{h=1}^{H} \sum_{i=1}^N  \frac{e^{\vz^T \mW_{1,h}^T\mW_{2,h} \vh_i/T}}{Z_h}  \frac{\eta}{H} \mW_{1,h}^T\mW_{2,h} \vh_i 
	\end{aligned} 
\end{equation*}
where $Z_h =  \sum_{j=1}^N  e^{\vz^T \mW_{1,h}^T\mW_{2,h} \vh_j/T}$.
Comparing with Eq~(\ref{eq:mha}), we can set $\mW_{1,h}^T\mW_{2,h} = \mW_{Q,h}^T\mW_{K,h} $ and $\mW_{O,h}\mW_{V,h}= \frac{\eta}{H} \mW_{1,h}^T \mW_{2,h}$ for $h \in [H]$. Then, we will have $Z'_h = Z_h$ and ${\rm MHA}(\vz) = \hat{\vz}$.
The remaining discussion is similar to that of single-head case in Section \ref{sec:all-framework}.

In Section~\ref{sec:all-framework}, we have shown that the forward pass of the attention model can be viewed as optimizing the global energy $F$ by one step first-order GD.
Then a natural idea arises: can we directly obtain the final solution as the token output (i.e., $\vz^* = {\rm argmin}_{\vz}F$) instead of relying on such an incremental structure driven by local GD?
Unfortunately, except in certain special cases (e.g., $\vh_i$ are symmetrically distributed), it is difficult to directly obtain a closed-form analytical solution for $F$ or its upper bound $\tilde{F}$. 
We present Lemma~\ref{app:lemma:mha-minima} as follows.
\begin{lemma}[Informal]\label{lemma:mha-minima}
	Both the global energy $F$ and its upper bound $\tilde{F}$ are non-convex with respect to $\vz$.
	Assume $\|\mW_{1,h}\vz\| \le \rho$ and $\| \mW_{2,h}\vh_i \| \le \rho$ for all $i \in [N]$ and $h \in [H]$, the local minima of $F$ are attained at the boundary or the stationary points. In addition, the local minima of $\tilde{F}$ can only be attained at the boundary.
\end{lemma}
The detailed proof of Lemma \ref{lemma:mha-minima} is in Appendix \ref{appsec:lemma-minima}.
The core is to show the Hessian matrix of $F$ as
\begin{equation}\label{eq:hessian}
	\nabla^2_{\vz} F
	=
	\frac{1}{H}\sum_{h=1}^H
	\Bigg[
	\underbrace{\vphantom{\frac{1}{T}}\mW_{1,h}^T \mW_{1,h}}_{\succeq 0}
	\underbrace{- \frac{1}{T}\mC_h}_{\preceq 0}
	\Bigg],
\end{equation}
which is composed of a positive semidefinite identity matrix and a negative semidefinite term for each head.
Therefore, $F$ is neither convex nor concave and its local minima can only occur at the boundary or at stationary points.
Similarly, the Hessian of $\tilde{F}$ contains only the negative semi-definite term, making it concave and ensuring that its local minima occur only on the boundary.
Although a closed-form solution is difficult to obtain directly, it is possible to obtain a better solution as the token representation by adopting more efficient GD algorithms within the energy-based framework, which in turn leads to improvements in the attention structure.
In the following, we design modifications to the attention mechanism based on first-order and second-order gradient descent algorithms respectively.

\begin{figure*}[t]
	\centering
	\includegraphics[scale=1.20]{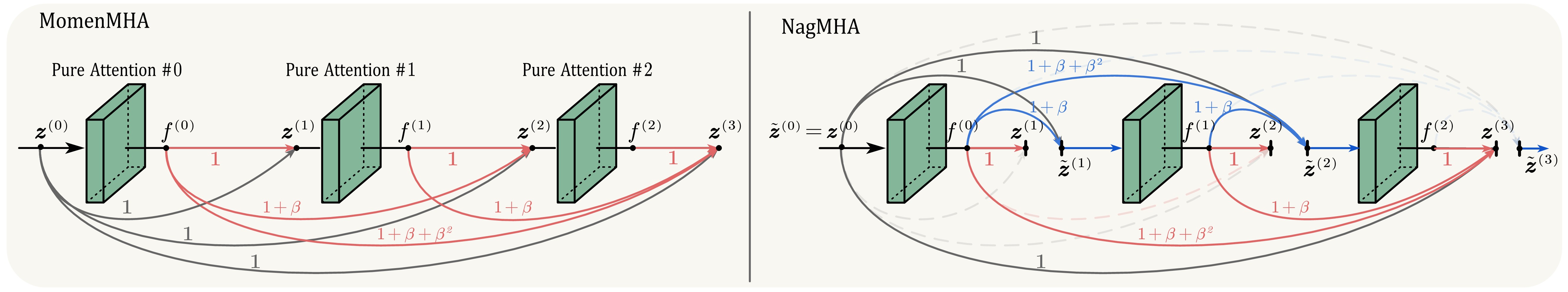}
	\caption{
		The illustration of the skip connections induced by ${\rm MomenMHA}$ and ${\rm NagMHA}$. We show the computation of $\vz^{(L)}$ in the case $L=3$ and $\eta = 1$. 
		The output $\vz^{(k)}$ of the $k$-th layer can be viewed as a weighted sum of the initial input $\vz^{(0)}$ (shown in gray) and the pure attention outputs $f^{(k)}$ from the previous $k-1$ layers (shown in pink). 
		The weight of each skip connection is indicated next to it.
		As for ${\rm NagMHA}$, the lookahead output $\tilde{\vz}^{(k)}$ forms upper-side skip connections (shown in blue).
		By introducing $\tilde{\vz}^{(k)}$, additional nodes are incorporated into the information flow, resulting in denser skip connections compared to  ${\rm MomenMHA}$.
	}
	\label{fig:test}
\end{figure*}

\subsection{Modifications based on 1st-order GD}\label{sec:atten1st}
The original softmax attention in Eq~(\ref{eq:mha}) can be viewed as performing one step first-order GD, i.e.,
\begin{equation}\label{eq:GD}
	\vz^{(k+1)} ={\rm MHA}(\vz^{(k)}) = \vz^{(k)} + f^{(k)} = \vz^{(k)} - \eta \nabla_{\vz^{(k)}} F,
\end{equation}
where we denote the pure attention update part (without skip connection) as $f^{(k)}$, which models the gradient part $f^{(k)} = - \nabla_{\vz^{(k)}} F$ with $\eta = 1$.
Considering the extensive literatures on GD, we can readily draw inspiration from them to inform modifications.
The momentum-based GD algorithm \cite{NAG} can be written as\footnote{It should be noted that here we put the learning rate $\eta$ in the update of $\vz$, which is slightly different from Sutskever \textit{et al.}~\cite{NAG}, where $\eta$ appears in front of the gradient. A similar modification is also applied in the NAG formulation. This form is closer to a PyTorch-style optimizer (torch.optim.SGD) implementation.}
\begin{align}\label{eq:momenGD}
	\left\{
	\begin{aligned}
		\vp^{(k+1)} &= \beta \vp^{(k)} + \nabla_{\vz^{(k)}} F ,\\
		\vz^{(k+1)} &= \vz^{(k)} - \eta \vp^{(k+1)} ,
	\end{aligned}
	\right.
\end{align}
where $\vp$ denotes the momentum and is initialized as $\vp^{(0)} = \vzero$, $\beta$ is a coefficient controlling the decay of past gradients, and $\eta$ is the learning rate.
By comparing with the original Eq~(\ref{eq:GD}), we can find that it suffices to replace $ - \nabla_{\vz^{(k)}} F$ in Eq~(\ref{eq:momenGD}) with the update part $f^{(k)}$ in Eq~(\ref{eq:GD}). 
We set $\beta$ as a two-layer input-dependent gate $\beta(\vz)$ while $\eta$ as learnable parameters initialized to $1.0$ respectively.
We refer to this modified structure as ${\rm MomenMHA}$.
Intuitively, the original attention mechanism can be viewed as modeling the gradient updates directly, whereas ${\rm MomenMHA}$ can be seen as learning the momentum updates.
In fact, we can expand the update of $\vz$ in Eq (\ref{eq:momenGD}) to obtain that  
$$
\vz^{(L)} = \vz^{(0)} + \sum_{k=0}^{L-1} \left(\sum_{n=0}^{L-1-k} \eta \beta^n \right) f^{(k)}.
$$
We can see that the output of the $L$-th layer can be viewed as a weighted sum of the initial input $\vz^{(0)}$ and the pure attention outputs $f^{(k)}$ from all previous $L-1$ layers with weights following a geometric progression of $\beta$.
This can be seen in Figure \ref{fig:test}.
When $\beta = 0$, it degenerates to the case of standard residual connection.

Compared with the original attention form, ${\rm MomenMHA}$ needs to maintain a momentum vector $\vp$ with the same shape as the input during the forward computation, thus introducing little additional overhead.
In fact, prior work \cite{sander2021momentum} has shown that maintaining such an additional momentum vector can help reduce memory storage during backpropagation by reconstructing activations on the fly, thereby offering potential savings in storage. 
We leave this implementation to future work.
In addition, to enable the momentum $\vp$ to propagate through the Feed-Forward Network (FFN) layers as well, we also replace $-\nabla_{\vz^{(k)}} F$ with the output of FFN modules similarly.

Another GD variant to further accelerate the convergence is using Nesterov Accelerated Gradient (NAG) \cite{NAG}, which introduces a lookahead mechanism that estimates the future position before computing the gradient.
This can be formalized as
\begin{align}\label{eq:nagGD}
	\left\{
	\begin{aligned}
		&\tilde{\vz}^{(k)} = \vz^{(k)} - \beta \vp^{(k)}, \\
		&\vp^{(k+1)} = \beta \vp^{(k)} + \nabla_{\tilde{\vz}^{(k)}} F ,\\
		&\vz^{(k+1)} = \vz^{(k)} - \eta \vp^{(k+1)} ,
	\end{aligned}
	\right.
\end{align}
where $\tilde{\vz}^{(k)}$ denotes the lookahead (or predicted) position obtained by moving along the momentum direction.
Similar to ${\rm MomenMHA}$, we just need to replace $- \nabla_{\tilde{\vz}^{(k)}} F$ with the update part $f^{(k)}$ in Eq~(\ref{eq:GD}) and we call this modification as ${\rm NagMHA}$.
We can also expand $\tilde{\vz}$ and $\vz$ to obtain the relationship among skip connections, as shown in Figure \ref{fig:nag-simple}.
In fact, by incorporating lookahead representations $\tilde{\vz}^{(k)}$ as new nodes, ${\rm NagMHA}$ introduces denser skip connections compared to  ${\rm MomenMHA}$. 
We defer the detailed discussion to the Appendix~\ref{app:MomenMHA}.
In addition, the remaining design details and discussions are similar to those in ${\rm MomenMHA}$, and are thus omitted for brevity.

\subsection{Modifications based on Newton's Method}\label{sec:atten2nd}
In addition to first-order momentum-based methods, another simple and straightforward idea for employing a more efficient algorithm is to use Newton’s method, which can be formulated as
\begin{equation*}
	\vz^{(k+1)} = \vz^{(k)} - \eta  \left[ \nabla_{\vz^{(k)}}^2 F \right]^{-1}  \nabla_{\vz^{(k)}} F,
\end{equation*}
where $\nabla_{\vz^{(k)}}^2 F$ is the Hessian matrix at $\vz^{(k)}$.
The above update can be viewed as preconditioning the gradient with the second-order information from the Hessian matrix.

To identify the induced second-order attention structure before introducing learnable projections, we first consider an unparameterized subspace.
Let $\vz^{(h)},\vu_i^{(h)}\in\mathbb{R}^{d_h}$ denote the representations of the current token and the interacting token $\vh_i$ in the $h$-th subspace respectively.
We denote the energy in the $h$-th subspace as $F_h= -T\log\sum_{i=1}^{N} \exp\left( (\vz^{(h)})^T\mathbf{u}_i^{(h)} / T \right)$ and then the global energy is $F = \frac{1}{H}\sum_{h=1}^H F_h$.
Instead of applying Newton’s method directly to $F$, we apply it independently to each subspace $F_h$, that is,
\begin{equation*}
	\vz^{(k+1)} = \vz^{(k)} - \frac{\eta}{H}\sum_{h=1}^H \left[ \nabla_{\vz^{(k)}}^2 F_h \right]^{-1}  \nabla_{\vz^{(k)}} F_h.
\end{equation*}

Denoting $\bar{\mathbf{u}}^{(h)} = \sum_i p_{i,h}\mathbf{u}_i^{(h)}$ where the normalized weight is $p_{i,h} = \operatorname{softmax}_i \big((\vz^{(h)})^T\mathbf{u}_i^{(h)}/T\big)$,
we have $\nabla_{\vz^{(h)}}F_h=-\bar{\mathbf{u}}^{(h)}$ and the Hessian matrix is
\begin{equation*}
    \nabla_{\vz^{(h)}}^2F_h = -\frac{1}{T} \sum_{i=1}^{N} p_{i,h}
	\left( \mathbf{u}_i^{(h)}-\bar{\mathbf{u}}^{(h)} \right) 
    \left( \mathbf{u}_i^{(h)}-\bar{\mathbf{u}}^{(h)} \right)^T.
\end{equation*}
For notation simplify, we denote the attention-weighted covariance part as $\mC_h = - T \nabla_{\vz^{(h)}}^2F_h \in\mathbb{R}^{d_h\times d_h}$.
It can be seen that the key to incorporating the second-order information is to use the curvature encoded by $\mC_h$ to precondition the first-order update.
Noticing that the global sign factors can be absorbed into the subsequent learnable projections, we therefore use $\mC_h^{-1}$ as the basic form of the preconditioner.

To further improve invertibility and numerical stability~\cite{bonnans2006numerical}, we introduce an identity damping term and formulate the preconditioner as
\begin{equation*}
\mP_h = \left( \mI+\alpha_h\mC_h \right)^{-1},
\end{equation*}
where $\alpha_h>0$ controls the strength of the curvature-dependent correction.
In addition, noticing that directly constructing and inverting $\mC_h$ for every query introduces considerable overhead, therefore we approximate it using only its diagonal entries~\cite{saad2003iterative}, that is, 
\begin{equation*} 
    \mC_h \approx \operatorname{Diag}(\vs_h),
\end{equation*}
\begin{equation*}
	\vs_h = \sum_{i=1}^{N} p_{i,h} \left( \vu_i^{(h)} \odot \vu_i^{(h)} \right) - \bar{\vu}^{(h)} \odot \bar{\vu}^{(h)}.
\end{equation*}
Here $\vs_h$ is exactly the coordinate-wise variance.
Compared with the full covariance matrix, this diagonal approximation retains the curvature magnitude along each feature dimension while discarding the cross-dimensional correlations.
More importantly, its inverse can be evaluated element-wise without explicitly constructing or inverting a matrix.

We next instantiate the above second-order structure into the attention modification.
As in standard attention, for the $h$-th head, let
$\vq_h=\mW_{Q,h}\vz$, $\vk_{i,h}=\mW_{K,h}\vh_i$, and $\vv_{i,h}=\mW_{V,h}\vh_i$,
with attention weights $p_{i,h}=\operatorname{softmax}_i(\vq_h^T\vk_{i,h}/T)$.
The first-order update is $\bar{\vv}_h=\sum_i p_{i,h}\vv_{i,h}$, and the corresponding diagonal second-order statistic is
\begin{equation*}
	\vs_h = \sum_{i=1}^{N} p_{i,h} \left(
	\vv_{i,h}\odot\vv_{i,h} \right) - \bar{\vv}_h\odot\bar{\vv}_h .
\end{equation*}
Here, the query and key projections determine the attention weights, while the value projection defines the space in which both the update and its curvature are evaluated.
In addition, to stabilize the scale of the estimated curvature, we normalize $\vs_h$ by its mean within each head as $\widetilde{\vs}_h=\operatorname{sg}(\vs_h)/(\operatorname{mean}(\operatorname{sg}(\vs_h))+\epsilon)$ where $\operatorname{sg}(\cdot)$ denotes the stop-gradient operator.
We further use a token-dependent two-layer gate $\alpha_h(\vz)>0$ to adaptively control the strength of the second-order correction for each head.
Combining the above components, we obtain our final attention modification based on second-order information as
\begin{equation*}
	\begin{aligned}
		{\rm DiagMHA2nd}(\vz) &= \vz + \sum_{h=1}^{H} \mW_{O,h} \mD_h \bar{\vv}_h,
		\\
		\mD_h &=\left[\mI+\alpha_h(\vz)\operatorname{Diag}\left(\widetilde{\vs}_h\right)\right]^{-1}.
	\end{aligned}
\end{equation*}
Noticing that $\mD_h$ is diagonal,  the preconditioning can be implemented element-wise as
$\mD_h\bar{\vv}_h = \bar{\vv}_h \oslash \big( 1+\alpha_h(\vz)\widetilde{\vs}_h \big)$,
where $\oslash$ denotes element-wise division.
Compared with standard MHA, the main additional computation is the attention-weighted second moment $\sum_i p_{i,h}(\vv_{i,h}\odot\vv_{i,h})$, which has the same asymptotic complexity as the standard value aggregation $\sum_i p_{i,h}\vv_{i,h}$.
The remaining operations only involve element-wise computation and a lightweight scalar gate, whose costs are lower order.
Therefore, DiagMHA2nd preserves the same asymptotic computational complexity as standard attention.
More detailed discussion can be seen in Appendix~\ref{app:diag2nd}.

\section{Experiments}\label{sec:ex}

\begin{figure*}[t]
    \centering
    \begin{minipage}[t]{0.49\textwidth}
        \centering
        \includegraphics[width=\linewidth]
        {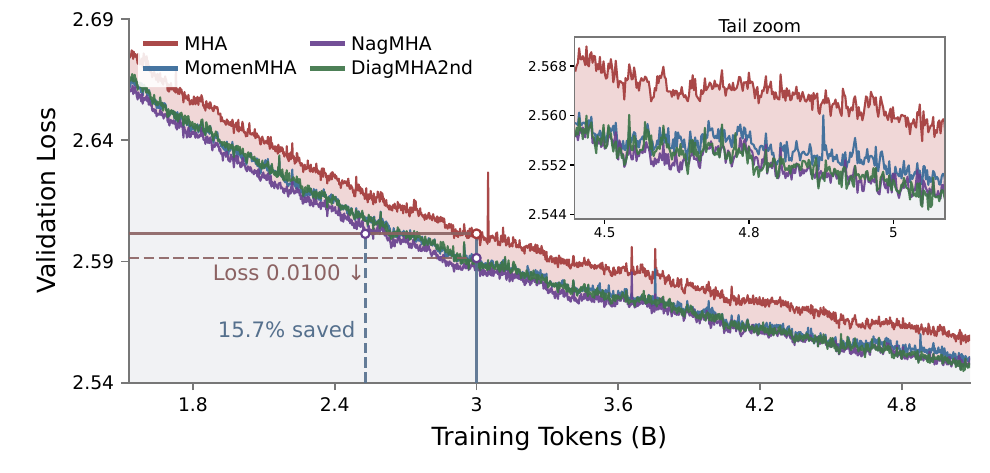}
        
        \vspace{-2mm}
        {\small (a) 160M - Training Tokens}
    \end{minipage}
    \hfill
    \begin{minipage}[t]{0.49\textwidth}
        \centering
        \includegraphics[width=\linewidth]
        {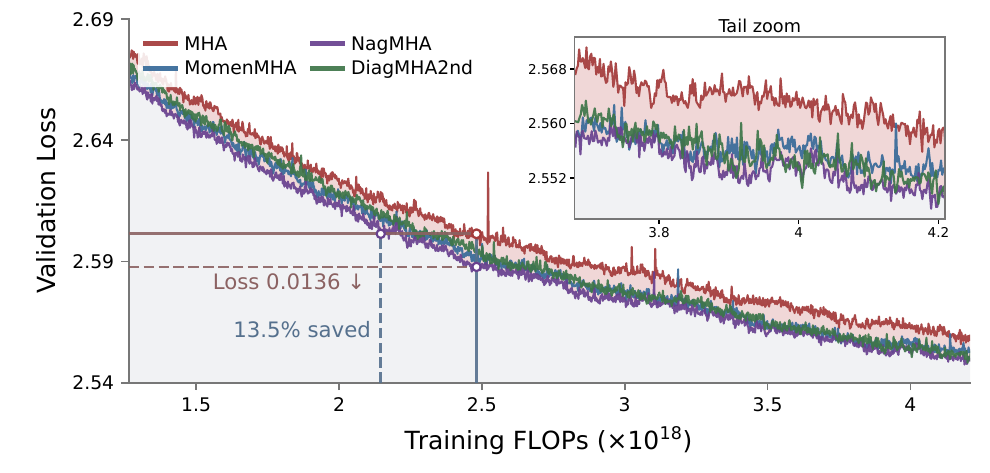}
        
        \vspace{-2mm}
        {\small (b) 160M - Training FLOPs}
    \end{minipage}
    \vspace{1mm}
    \begin{minipage}[t]{0.49\textwidth}
        \centering
        \includegraphics[width=\linewidth]
        {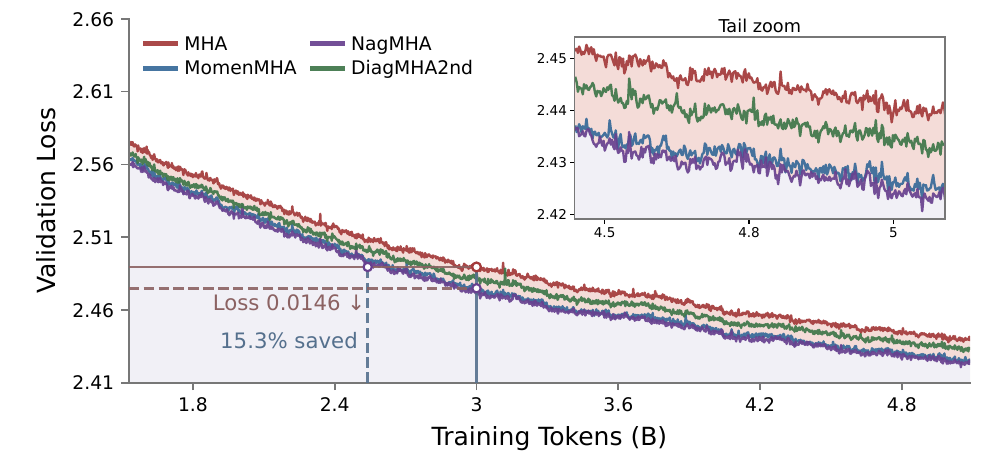}
        
        \vspace{-2mm}
        {\small (c) 300M - Training Tokens}
    \end{minipage}
    \hfill
    \begin{minipage}[t]{0.49\textwidth}
        \centering
        \includegraphics[width=\linewidth]
        {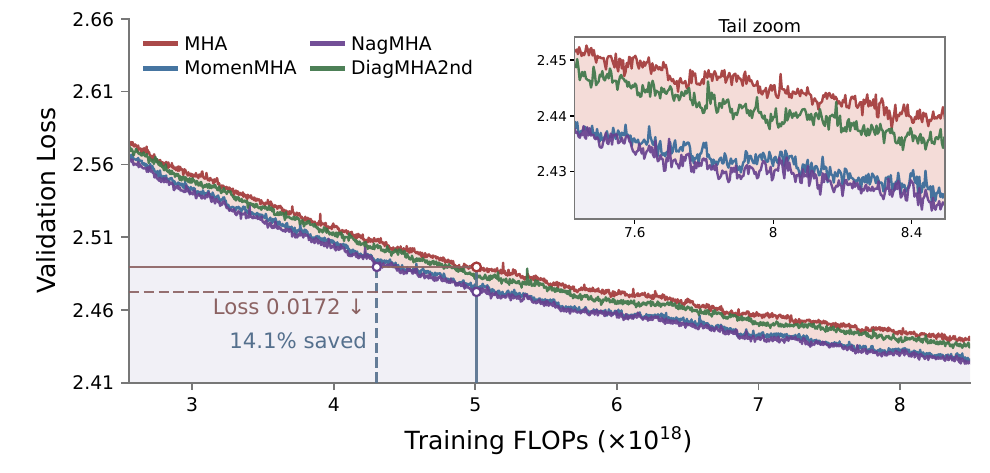}
        
        \vspace{-2mm}
        {\small (d) 300M - Training FLOPs}
    \end{minipage}
    \vspace{-0mm}
    \caption{
        \textbf{Validation loss and training efficiency of different attention mechanisms at the 160M and 300M model scales.}
        (a,b) Validation loss as a function of cumulative training tokens and FLOPs for the 160M-scale models.
        (c,d) Corresponding results for the 300M-scale models.
        The annotations highlight the training cost savings achieved by NagMHA at matched validation loss and the loss reduction at matched training cost.
        The inset plots provide enlarged views of the final training stages.
    }
    \label{fig:pretrain}
    \vspace{-2mm}
\end{figure*}

\begin{figure}[t]
    \centering
    \begin{minipage}[t]{0.49\linewidth}
        \centering
        \includegraphics[width=\linewidth]{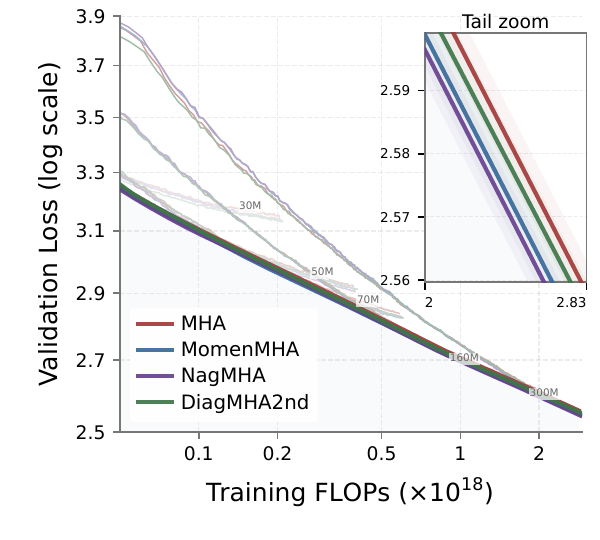}
    \end{minipage}
    \hfill
    \begin{minipage}[t]{0.49\linewidth}
        \centering
        \includegraphics[width=\linewidth]{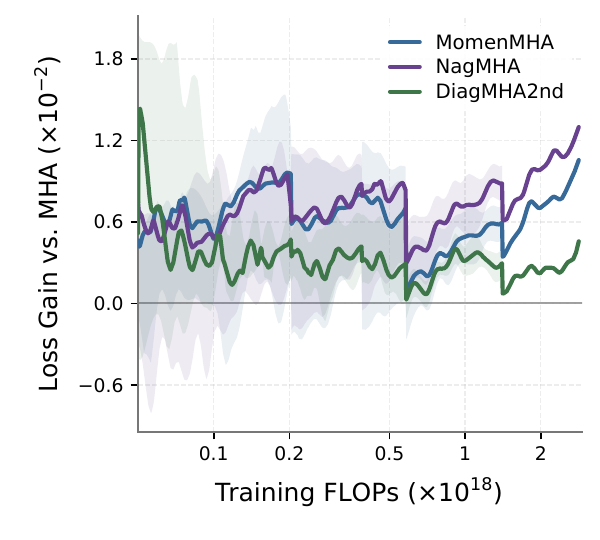}
    \end{minipage}
    \vspace{-6mm}
    \caption{
        \textbf{Compute--loss trends and relative improvements.}
        (a) Validation loss versus cumulative training FLOPs
        for different attention mechanisms and model sizes.
        Solid curves show the fitted lower envelopes.
        (b) Compute-matched loss improvements over MHA.
        Solid curves indicate the median improvement across
        available model sizes, with shaded regions showing
        the interquartile range (Q1--Q3).
    }
    \label{fig:scaling}
    \vspace{-0mm}
\end{figure}

\subsection{Basic Pretraining}

To explore the potential of the proposed attention modifications, we conduct experiments using a GPT-like architecture \cite{GPT2}. 
Specifically, we replace the original standard softmax attention with the ${\rm MomenMHA}$ and ${\rm NagMHA}$ introduced in Section \ref{sec:atten1st}, as well as the ${\rm DiagMHA2nd}$ described in Section \ref{sec:atten2nd}.
We conduct small-scale pre-training on the MiniPile dataset \cite{minipile}, which is a compact subset version of the original Pile dataset \cite{Pile}.
We use the GPT-2 tokenizer from huggingface \cite{huggingface} to process the corpus and the model sizes range from 30M to 300M parameters.
We conduct training on the training set containing 1 million samples with the objective of next-token prediction, while simultaneously monitoring and reporting the loss on the validation set~\footnote{Our code is available at {https://github.com/miaomouse/submit-code-energy-attn}}.
More details can be seen in Appendix~\ref{app:Ex}.

\definecolor{basegray}{gray}{0.92}
\definecolor{exampletitle}{HTML}{EEE9F7}

\newcolumntype{Y}{>{\centering\arraybackslash}X}

\begin{table*}[t]
\centering
\caption{
Pretraining and downstream evaluation results of different
attention mechanisms at 160M and 300M model scales.
All models are evaluated after three training epochs.
Accuracy metrics are reported as percentages.
{\bf Bold} and \underline{underlined} values indicate
the best and second-best results among individual models
in each column, respectively.
Rel. Gain denotes the relative improvement of NagMHA over the
corresponding MHA baseline.
}
\label{tab:attention_downstream}
\small
\setlength{\tabcolsep}{2.5pt}
\renewcommand{\arraystretch}{1.15}

\begin{tabularx}{\textwidth}{@{}l*{11}{Y}@{}}
\toprule
\multirow{2}{*}{Methods}
& \multicolumn{2}{c}{MiniPile}
& \multicolumn{2}{c}{WikiText}
& \multicolumn{3}{c}{LAMBADA}
& \multicolumn{1}{c}{HellaSwag}
& \multicolumn{1}{c}{PIQA}
& \multicolumn{1}{c}{ARC-Easy}
& \multicolumn{1}{c}{OpenBookQA} \\
\cmidrule(lr){2-3}
\cmidrule(lr){4-5}
\cmidrule(lr){6-8}
\cmidrule(lr){9-9}
\cmidrule(lr){10-10}
\cmidrule(lr){11-11}
\cmidrule(lr){12-12}
& NLL$\downarrow$ & PPL$\downarrow$
& NLL$\downarrow$ & PPL$\downarrow$
& Acc$\uparrow$ & NLL$\downarrow$ & PPL$\downarrow$
& Acc$_{\rm norm}\uparrow$
& Acc$_{\rm norm}\uparrow$
& Acc$_{\rm norm}\uparrow$
& Acc$_{\rm norm}\uparrow$ \\
\midrule

\rowcolor{basegray}
\multicolumn{12}{c}{{\it 160M Parameters}} \\

MHA
& 2.559 & 12.919
& 3.702 & 40.536
& 21.54 & 3.902 & 49.478
& 27.23 & 58.87 & 36.91 & 24.80 \\
\midrule

DiagMHA2nd
& 2.547 & 12.774
& 3.694 & 40.197
& 20.28 & 3.918 & 50.281
& 26.90 & 59.41 & 36.74
& \underline{26.60} \\

MomenMHA
& 2.551 & 12.815
& 3.698 & 40.351
& 22.26 & 3.867 & 47.793
& 26.99 & 59.68 & 36.62 & 25.20 \\

NagMHA
& 2.551 & 12.818
& 3.687 & 39.906
& 22.51 & 3.740 & 42.106
& 27.08 & 60.55 & 36.32 & 25.60 \\

\midrule

{\it Rel. Gain}
& {\it +0.31\%} & {\it +0.78\%}
& {\it +0.42\%} & {\it +1.55\%}
& {\it +4.50\%} & {\it +4.14\%}
& {\it +14.90\%}
& {\it -0.55\%} & {\it +2.87\%}
& {\it -1.60\%} & {\it +3.23\%} \\

\midrule

\rowcolor{exampletitle}
\multicolumn{12}{c}{{\it 300M Parameters}} \\

MHA
& 2.440 & 11.478
& 3.508 & 33.372
& 24.49 & 3.517 & 33.667
& 28.29 & 61.21 & \textbf{39.35} & 26.00 \\
\midrule

DiagMHA2nd
& 2.434 & 11.409
& 3.511 & 33.498
& 23.33 & 3.535 & 34.308
& \underline{28.63} & 61.43 & 38.13
& \textbf{27.60} \\

MomenMHA
& \underline{2.426} & \underline{11.308}
& \underline{3.491} & \underline{32.809}
& \textbf{26.74} & \textbf{3.329}
& \textbf{27.914}
& 28.41 & \underline{61.53}
& \underline{39.06} & 26.40 \\

NagMHA
& \textbf{2.424} & \textbf{11.291}
& \textbf{3.490} & \textbf{32.775}
& \underline{25.25} & \underline{3.386}
& \underline{29.541}
& \textbf{28.91} & \textbf{62.24}
& 38.51 & 26.40 \\

\midrule

{\it Rel. Gain}
& {\it +0.67\%} & {\it +1.63\%}
& {\it +0.51\%} & {\it +1.79\%}
& {\it +3.09\%} & {\it +3.72\%}
& {\it +12.25\%}
& {\it +2.18\%} & {\it +1.69\%}
& {\it -2.14\%} & {\it +1.54\%} \\




\bottomrule
\end{tabularx}
\end{table*}

First, we illustrate the relationship between validation loss and cumulative training tokens and FLOPs at the 160M model scale in Figs.~\ref{fig:pretrain}(a) and (b) respectively.
At the 160M scale, all three attention variants consistently achieve lower validation loss than standard MHA under the same training token budget, with NagMHA exhibiting the largest improvement.
When taking MHA after processing 3B tokens as the reference point, NagMHA reduces validation loss by 0.0100 at the same token budget, while requiring 15.7\% fewer tokens to achieve the same validation loss.
To account for the slight differences in parameter counts across attention variants, we further compare their performance in terms of training FLOPs.
As shown in Fig.~\ref{fig:pretrain}(b), the improvements remain consistent under matched computational budgets.
Specifically, at the computational cost corresponding to 3B training tokens for MHA, NagMHA achieves a validation loss reduction of 0.0136, while requiring 13.5\% fewer FLOPs to reach the same validation loss.
When further scaling the model up to 300M parameters, the advantages of the attention variants remain evident, with all three variants outperforming standard MHA under matched training token or computation budgets and NagMHA achieving the lowest validation loss.
Using the same reference point of 3B training tokens as well, NagMHA reduces validation loss by 0.0146 and requires 15.3\% fewer tokens to reach the same validation loss as MHA.
These improvements also persist when accounting for computational cost, as illustrated in Fig.~\ref{fig:pretrain}(d).
Under the FLOP budget corresponding to 3B training tokens for MHA, NagMHA reduces validation loss by 0.0172 and achieves a 14.1\% reduction in FLOPs required to reach the same validation loss.
Overall, these results demonstrate that the proposed attention modifications can improve both validation performance and training efficiency across different model scales.

To further investigate the scaling behavior of different attention mechanisms, we extend our analysis to five model scales, ranging from 30M to 300M parameters.
As shown in Fig.~\ref{fig:scaling}(a), we compare the compute--loss trends of different attention mechanisms by constructing the fitted lower envelopes of validation loss across model sizes.
Overall, the three attention variants exhibit more favorable compute--loss trends than standard MHA, with MomenMHA and NagMHA showing more pronounced improvements.
These results suggest that the proposed attention modifications can achieve lower validation loss under comparable computational budgets across different model scales.
To further quantify these improvements, Fig.~\ref{fig:scaling}(b) presents the loss reduction relative to MHA at matched training FLOPs.
We observe that DiagMHA2nd achieves modest improvements over standard MHA, while MomenMHA and particularly NagMHA yield more pronounced gains.
Overall, these results provide further evidence that the proposed attention modifications can improve the compute efficiency of language model pretraining across different model scales.

\begin{figure}[h]
    \centering
    \begin{minipage}[t]{0.49\linewidth}
        \centering
        \includegraphics[width=\linewidth]
        {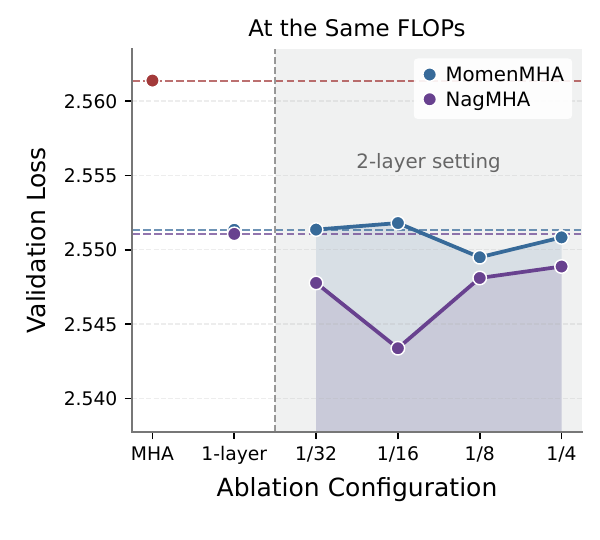}
    \end{minipage}
    \hfill
    \begin{minipage}[t]{0.49\linewidth}
        \centering
        \includegraphics[width=\linewidth]
        {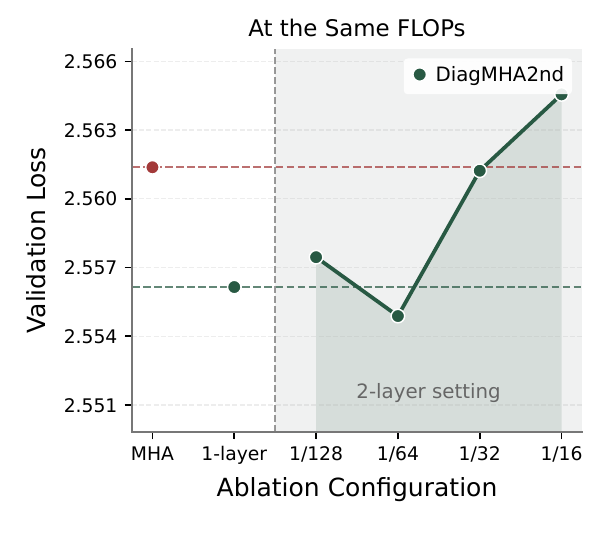}
    \end{minipage}
    \vspace{-2mm}
    \caption{
        \textbf{Ablation study on gating mechanisms.}
        Validation loss under matched training FLOPs after approximately one epoch of pretraining.
        (a) Comparison of different $\beta$-gate configurations in MomenMHA and NagMHA where the fractions denote the factors $m_{\beta}$ of the two-layer setting with $d_h=m_{\beta}d$.
        (b) Corresponding comparison for the $\alpha$-gate in DiagMHA2nd and the fractions denote the factor $m_{\alpha}$.
        The standard MHA baseline is included for reference.
    }
    \label{fig:gate_ablation}
    \vspace{-2mm}
\end{figure}

We further investigate the influence of the gating hyperparameters introduced in MomenMHA, NagMHA, and DiagMHA2nd.
Specifically, we compare different gate architectures and hidden dimensions under matched training FLOPs after approximately one epoch of pretraining.
For MomenMHA and NagMHA, we investigate the effect of the $\beta$-gate by comparing a single-layer gate with two-layer gates of different hidden dimensions $d_{h}=m_{\beta}d$, where $m_{\beta}\in\{\frac{1}{32},\frac{1}{16},\frac{1}{8},\frac{1}{4}\}$.
Specifically, for one-layer setting, the $\beta$-gate is formulated as $\beta(\vz)=\sigma\left(\vz^T\vw \right)$ where $\vw \in \sR^{d \times 1}$ and $d$ is the model dimension.
While the two-layer gate is $\beta(\vz)=\sigma\left(\operatorname{SiLU}\left(\vz^T\mA \right)\mB \right)$ where $\mA \in\mathbb{R}^{d\times m_{\beta}d}$, $\mB_h\in\mathbb{R}^{m_{\beta}d\times1}$ and $m_{\beta}$ is the factor.
As shown in the left part of Fig.~\ref{fig:gate_ablation}, the two-layer gates generally achieve more favorable validation loss than the single-layer setting, although the improvements depend on the hidden dimension.
In particular, MomenMHA achieves its lowest validation loss at $m_{\beta}=\frac{1}{8}$, while NagMHA performs best at $m_{\beta}=\frac{1}{16}$.
It can be seen that further increasing the hidden dimension does not consistently improve performance under the same computational budget.
These observations suggest that a moderate gate capacity is sufficient to capture useful adaptive information, while excessively increasing the gate size may instead diminish the gains under matched training FLOPs.
Similarly, we investigate the $\alpha$-gate in DiagMHA2nd by comparing a single-layer gate with two-layer gates of hidden dimensions $d_{h}=m_{\alpha}d$, where $m_{\alpha}\in\{\frac{1}{128},\frac{1}{64},\frac{1}{32},\frac{1}{16}\}$.
As illustrated in the right part of Fig.~\ref{fig:gate_ablation}, the two-layer gate with $m_{\alpha}=\frac{1}{64}$ achieves the lowest validation loss, outperforming the single-layer setting.
However, further increasing the hidden dimension to $m_{\alpha}=\frac{1}{32}$ or $m_{\alpha}=\frac{1}{16}$ leads to noticeable performance degradation under matched FLOPs.
These results indicate that employing a two-layer gate can further improve the performance over a single-layer gate while increasing the gate size does not necessarily yield better performance.
A more comprehensive comparison throughout pretraining can be seen in Fig.~\ref{fig:gate_ablation_dynamics} in the Appendix.

\subsection{Downstream Generalization}
Furthermore, we investigate the generalization capabilities of the pretrained model variants across a diverse set of evaluation benchmarks, including:
(a) \textbf{MiniPile}~\cite{minipile} and \textbf{WikiText}~\cite{WikiText}, which evaluate language modeling performance on held-out pretraining data and Wikipedia articles respectively.
We report negative log-likelihood (NLL) and perplexity (PPL) for both datasets;
(b) \textbf{LAMBADA}~\cite{paperno-etal-2016-lambada}, which evaluates the ability to predict the final word of a passage using broader discourse context.
We report prediction accuracy, target NLL, and target PPL;
(c) \textbf{HellaSwag}~\cite{zellers-etal-2019-hellaswag}, which evaluates commonsense reasoning by selecting the most plausible continuation of a given context;
(d) \textbf{PIQA}~\cite{PIQA}, which assesses physical commonsense reasoning;
(e) \textbf{ARC-Easy}~\cite{arc}, which consists of elementary-level multiple-choice science questions; and
(f) \textbf{OpenBookQA}~\cite{mihaylov-etal-2018-suit}, which evaluates scientific commonsense reasoning using elementary scientific facts.
For the multiple-choice benchmarks (c)--(f), we report length-normalized accuracy ($\mathrm{Acc}_{\mathrm{norm}}$).

The overall results are summarized in Table~\ref{tab:attention_downstream}.
Compared with standard MHA, DiagMHA2nd achieves modest improvements in MiniPile validation loss at both model scales, while its performance on downstream benchmarks remains mixed.
In contrast, MomenMHA and NagMHA demonstrate more pronounced improvements in language modeling and several downstream evaluations.
In particular, MomenMHA achieves strong performance on LAMBADA, reducing target PPL from 33.667 to 27.914 at the 300M scale, corresponding to a 17.09\% reduction.
Meanwhile, NagMHA exhibits favorable overall performance across both model scales, reducing WikiText PPL by 1.55\% and 1.79\% at 160M and 300M, respectively.
It also improves PIQA normalized accuracy from 58.87\% to 60.55\% and from 61.21\% to 62.24\%, while achieving the lowest MiniPile PPL and the highest HellaSwag normalized accuracy at the 300M scale.
However, the improvements on ARC-Easy and OpenBookQA are less consistent with those observed in language modeling, potentially due to the relatively small model sizes and distributional differences between the pretraining corpus and these downstream benchmarks.

\begin{figure}[t]
    \centering
    \begin{minipage}[t]{0.49\linewidth}
        \centering
        \includegraphics[width=\linewidth]
        {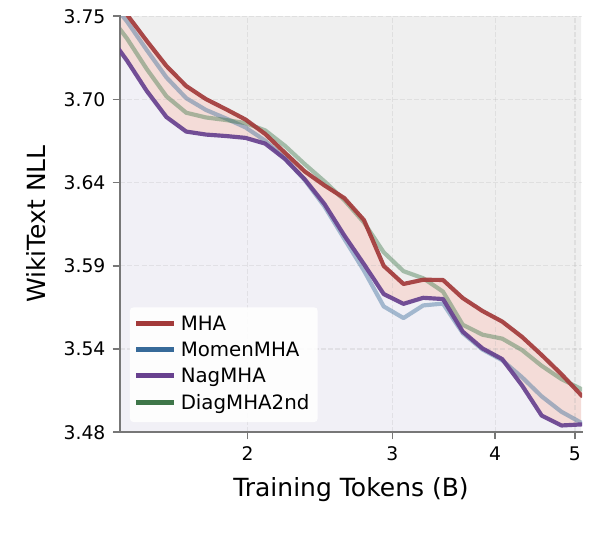}
        \vspace{-4mm}
        {\small (a) WikiText}
    \end{minipage}
    \hfill
    \begin{minipage}[t]{0.49\linewidth}
        \centering
        \includegraphics[width=\linewidth]
        {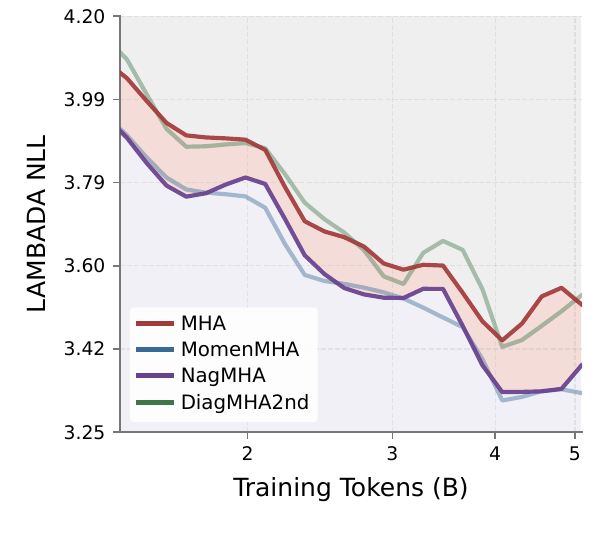}
        \vspace{-4mm}
        {\small (b) LAMBADA}
    \end{minipage}
    \vspace{-0mm}
    \caption{
        \textbf{Generalization dynamics during pretraining.}
        NLL on (a) WikiText and (b) LAMBADA, evaluated using intermediate checkpoints
        of the 300M-scale models.
        The curves illustrate the generalization performance of different attention mechanisms as pretraining progresses.
    }
    \label{fig:generalization_dynamics}
    \vspace{-2mm}
\end{figure}

In addition, we also evaluate intermediate checkpoints of the 300M models on WikiText and LAMBADA, as illustrated in Fig.~\ref{fig:generalization_dynamics}.
Overall, MomenMHA and NagMHA generally achieve lower NLL than standard MHA on both datasets throughout much of the training process, with particularly noticeable improvements on LAMBADA.
Although the evaluation curves exhibit some fluctuations, the advantages of these two variants remain evident in the later training stages.
These observations further suggest that the benefits of the attention modifications extend beyond the pretraining validation loss and can generalize to other language modeling tasks.

\section{Conclusion}
In this paper, we revisited attention-based models through the energy principle and presented a unified energy-based framework for understanding attention as an implicit optimization process. By specifying the local energy, global energy, and optimization algorithm, the framework recovers several representative attention forms, including linear attention, gated linear attention, and standard softmax attention. Building on this perspective, we further derived attention modifications inspired by momentum-based gradient descent, Nesterov accelerated gradient, and Newton's method. Experiments on language model pretraining across different model scales and downstream benchmarks provide preliminary evidence that the proposed modifications can improve training efficiency and generalization. We hope that this energy-based viewpoint can provide useful guidance for interpreting and designing future attention architectures.

\section*{Acknowledgments}
This research was supported in part by Beijing Natural Science Foundation under Grant Z250001, in part by National Key Research and Development Program of China under Grant 92570203 and Grant 2024YFE0203200, in part by National Natural Science Foundation of China under Grant 62476277, in part by CCF-ALIMAMA TECH Kangaroo Fund under Grant CCF-ALIMAMA OF 2024008, in part by Huawei-Renmin University joint program on Information Retrieval, in part by the fund for building worldclass universities (disciplines) of Renmin University of China, in part by the funds from Beijing Key Laboratory of Big Data Management and Analysis Methods, Gaoling School of Artificial Intelligence, Renmin University of China, in part by the Engineering Research Center of Next-Generation Intelligent Search and Recommendation, in part by the Ministry of Education, from Intelligent Social Governance Interdisciplinary Platform, in part by the Major Innovation \& Planning Interdisciplinary Platform for the ``DoubleFirst Class'' Initiative, in part by Renmin University of China, in part by the Public Policy and Decision-making Research Lab of Renmin University of China, and in part by the Public Computing Cloud of Renmin University of China.

\appendices
\section{Additional Discussions and Proofs}\label{app}
\subsection{Discussion on More Ingredients}\label{app:ingredients}


In Section~\ref{sec:all-framework}, we show how to make appropriate choices within the proposed framework and establish connections with existing major forms of attentions. 
In fact, modifications to the three key components in the existing scenario, namely local energy $E_i$, global energy $F$, and the optimization algorithm, can analogously lead to new forms of attention. 
Note that the three aspects mentioned above are relatively independent, meaning that different ingredients can be combined in various ways, leading to a rich variety of attention forms.
Here, we discuss more existing or potential ingredients:

The local energy in the existing scenario is directly taken in the form of a parameterized inner product or $\ell_2$ regression loss. Both can be viewed as a measure of similarity. 
A natural idea is to use a more general $\ell_p$ norm-based loss, that is, $E_i = \| \vz - \mW \vh\|_p^p$ where $p \ge 1$ and $\| \cdot \|_p$ is the $\ell_p$-norm for vectors.
We may want to use different values of $p$ depending on the data and scenario to achieve the desired properties.
Additionally, another approach is that although the $\ell_2$-norm is a natural choice in many machine learning tasks, it is sensitive to outliers and extreme values. 
Therefore, it can be further extended to a more robust Huber loss-type \cite{huber1992robust, hastie2009elements, miras}, that is,
\begin{equation*}
	E_i = E(\vz, \vh_i) = \left\{ \begin{matrix}
		\frac{1}{2} \|\vz - \mW\vh_i \|^2 &  \| \vz - \mW\vh_i\| \le \delta \\
		\delta \left( \| \vz - \mW \vh_i\| - \frac{1}{2} \delta \right)	& \| \vz - \mW\vh_i\| > \delta
	\end{matrix}\right.
\end{equation*}
where $\delta$ is a threshold.
This loss function reduces the impact of outliers by applying a linear loss for large deviations.
We leave it to further exploration whether these potential choices could enhance the attention structure in certain situations.

For global energy, existing literatures on sparse modern Hopfield networks \cite{hu2023sparse, wu2023stanhop} have established a generalized sparse Hopfield energy function, where Tsallis $\alpha$-entropies \cite{tsallis1988possible} are used to yield retrieval dynamics corresponding to various sparse attentions.
Softmax attention and sparsemax attention can be viewed as special cases with $\alpha = 1$ and $\alpha = 2$ respectively \cite{martins2016softmax, peters2019sparse}.
More general values of $\alpha$ can lead to attention structures with varying degrees of sparsity \cite{correia2019adaptively}.
In addition, norm $\gamma$-negentropies can also be used to derive attention patterns different from the sparse attention modes above, where $\gamma = 1$ corresponds to agrmax, while $\gamma \rightarrow +\infty$ encourages sparse and uniform attention scores \cite{blondel2020learning, santos2024sparse}.

As for the optimization algorithm, one-step first-order gradient descent is mainly used in the recipes of the discussed different attention forms.
In fact, since there are lots of mature works in optimization literatures, it is possible for us to achieve this by drawing
on these works.
We explore this in Section \ref{sec:modification} to induce new forms of attention.

\subsection{Proof of Lemma \ref{app:lemma:mha-minima}}\label{appsec:lemma-minima}

Before presenting the proof of Lemma \ref{app:lemma:mha-minima} for the multi-head case, we first provide the analysis for the single-head scenario as follows.
We define $\tilde{F} = -T \log \sum_{i=1}^{N} e^{\frac{\vz^T \mW\vh_i}{T}}$.
When we relax the constraint for $E_i$ in the softmax attention recipe so that $\|\vz\| \le \rho$ and $\| \mW\vh_i \| \le \rho $ for all $i \in [N]$, we will have $F = - T \log \sum_{i=1}^N  e^{- \frac{\| \vz - \mW \vh_i\|^2}{2T}} \le \tilde{F} + \rho^2$.
For simplicity, we refer to $\tilde{F}$ as the upper bound of $F$ despite differing by a constant $\rho^2$.
Next we present the lemma for the single-head case.
\begin{lemma}[Single-head Case]\label{lemma:minima}
	Assuming $\|\vz\| \le \rho$ and $\| \mW\vh_i \| \le \rho $ for all $i \in [N]$, both the global energy $F$ and its upper bound $\tilde{F}$ are non-convex with respect to $\vz$.
	The local minima of $F$ are attained at the boundary $\|\vz\| = \rho$ or when $\vz = \sum_{i=1}^N p_i  \mW \vh_i$ where $p_i = \frac{1}{Z}e^{- \frac{\| \vz - \mW\vh_i \|^2}{2T}}$ and  $Z = \sum_{i=1}^{N} e^{- \frac{\| \vz - \mW\vh_i \|^2}{2T}}$. In addition, the local minima of $\tilde{F}$ are attained at the boundary $\|\vz\| = \rho$.
\end{lemma}
\begin{proof}
	Recalling that in the single-head case $F = - T \log \sum_{i=1}^N  e^{- \frac{\| \vz - \mW \vh_i\|^2}{2T}}$. We can compute the derivative of $F$ with respect to $\vz$ as
	\begin{equation*}
		\nabla_{\vz} F =  -T  \nabla_{\vz} \log \sum_{i=1}^{N} e^{- \frac{\|\vz - \mW \vh_i\|^2}{2T}} = \sum_{i=1}^N p_i \left( \vz - \mW \vh_i \right),
	\end{equation*}
	where $p_i = \frac{1}{Z}e^{- \frac{\| \vz - \mW\vh_i \|^2}{2T}}$ and $Z = \sum_{i=1}^{N} e^{- \frac{\| \vz - \mW\vh_i \|^2}{2T}}$.
	For notational simplicity, we denote $\vr_i = \vz - \mW \vh_i$.
	To compute the Hessian matrix, we first calculate
	\begin{equation*}
		\begin{aligned}
			\nabla_{\vz} p_i &= \nabla_{\vz} \frac{e^{- \frac{\| \vr_i \|^2}{2T}}}{Z} = - \frac{\vr_i}{TZ} e^{ - \frac{\| \vr_i \|^2}{2T}} +  \frac{e^{- \frac{\| \vr_i \|^2}{2T}} }{TZ^2} \sum_{j=1}^{N} \vr_j  e^{-\frac{\| \vr_j \|^2}{2T}} \\
			&= - \frac{1}{T} p_i \vr_i + \frac{1}{T} p_i \sum_{j=1}^N p_j \vr_j  
		\end{aligned}
	\end{equation*}
	Therefore, the Hessian matrix of $F$ with respect to $\vz$ is
	\begin{equation*}
		\begin{aligned}
			\nabla^2_{\vz} F &= \sum_{i=1}^{N} \vr_i \left( - \frac{1}{T} p_i \vr_i^T + \frac{1}{T} p_i \sum_{j=1}^N p_j \vr_j^T  \right) + \mI \\
			&= \mI - \frac{1}{T} \sum_{i=1}^N p_i \vr_i \vr_i^T + \frac{1}{T} \sum_{i=1}^N p_i \vr_i \sum_{j=1}^N p_j \vr_j^T \\
			&= \mI - \frac{1}{T} \left[ \sum_{i=1}^N p_i \vr_i \vr_i^T -  \left(\nabla_{\vz} F\right) \left(\nabla_{\vz} F \right)^T  \right]. 
		\end{aligned}
	\end{equation*}
	Furthermore, for any $\vv \in \sR^d$, we have
	\begin{equation}\label{eq:temphessian}
	\begin{aligned}
		\vv^T \nabla^2_{\vz} F \vv = \| \vv \|^2 - \frac{1}{T} \left[ \sum_{i=1} p_i \vv_i^T \vr_i \vr_i^T \vv_i - \left( \vv^T \nabla_{\vz} F \right) \left( \vv^T  F\right)^T \right]
	\end{aligned}
	\end{equation}
	Let $X_i = \vr_i^T \vv$ and define a random variable $X$ such that $P(X = X_i) = p_i$. Then for the second term in Eq~(\ref{eq:temphessian}), we have 
	\begin{equation*}
	\begin{aligned}
		&- \frac{1}{T} \left[ \sum_{i=1}^N p_i \| \vr_i^T \vv \|^2  - \left( \sum_{i=1}^N p_i \vr_i^T \vv \right)^2  \right] \\
		&= -\frac{1}{T} \left[  \E\left(X_i^2\right) - \E^2\left(  X_i \right) \right] \\
		&= - \frac{1}{T} \Var(X) \le 0.
	\end{aligned}
	\end{equation*}
	Considering that the identity matrix is positive semi-definite, we obtain
	\begin{equation*}
		\nabla^2_{\vz} F 
		= \underbrace{\vphantom{\Bigg(}\mI}_{\succeq 0} 
		\underbrace{ - \frac{1}{T} \left[ \sum_{i=1}^N p_i \vr_i \vr_i^T 
			- (\nabla_{\vz} F)(\nabla_{\vz} F)^T \right]}_{\preceq 0}.
	\end{equation*}
	Therefore, we obtain that $F$ is neither convex nor concave and when $\| \vz \| \le \rho$, its local minima can only be attained at the boundary $\| \vz \| = \rho $ or at interior points where $\nabla_{\vz} F = 0$, that is, $\vz = \sum_{i=1}^N p_i \mW \vh_i$.
	
	Similarly, we can obtain the Hessian matrix of $\tilde{F}$ as
	\begin{equation*}
		\nabla^2_{\vz} \tilde{F} =  - \frac{1}{T} \left[ \sum_{i=1}^N p_i (\mW\vh_i) (\mW\vh_i)^T 
		- (\nabla_{\vz} \tilde{F})(\nabla_{\vz} \tilde{F})^T \right] \preceq 0,
	\end{equation*}
	where $p_i = \frac{e^{\vz^T \mW \vh_i / T}}{Z}$ and $Z = \sum_{i=1}^N e^{\frac{\vz^T\mW\vh_i}{T}}$. Therefore, we can get that $\tilde{F}$ is concave and when $\| \vz \| \le \rho$, its local minima can only be attained at the boundary $\| \vz \| = \rho $.
\end{proof}

We now present the proof of Lemma \ref{app:lemma:mha-minima}.
In fact, noting that each head is independent, the proof is very similar to that of the single-head case.
We define $\tilde{F} = - \frac{1}{H} \sum_{h=1}^{H} T \log \sum_{i=1}^{N} e^{-  \vz^T \mW_{1,h}^T \mW_{2,h} \vh_i / T}$.
Similarly, when we relax the constraint for $E_{h,i}$ in the recipe so that $\|\mW_{1,h}\vz\| \le \rho$ and $\| \mW_{2,h}\vh_i \| \le \rho $ for all $i \in [N]$ and $h \in [H]$, we will have $F \le \tilde{F} + \rho^2$.
For simplicity, we also refer to $\tilde{F}$ as the upper bound of $F$ despite differing by a constant $\rho^2$.
Next we present the lemma for the multi-head case.
\begin{lemma}\label{app:lemma:mha-minima}
	Both the global energy $F$ and its upper bound $\tilde{F}$ are non-convex with respect to $\vz$. 
	Assuming $\|\mW_{1,h}\vz\| \le \rho$ and $\| \mW_{2,h}\vh_i \| \le \rho$ for all $i \in [N]$ and $h \in [H]$, the local minima of $F$ are attained at the boundary $\|\mW_{1,h}\vz\| = \rho$ or when $\sum_{h=1}^{H}\sum_{i=1}^N p_{i,h} \mW_{1,h}^T \left( \mW_{1,h} \vz - \mW_{2,h} \vh_i \right) = 0$ where $p_{i,h} = \frac{1}{Z_h}e^{- \| \mW_{1,h} \vz - \mW_{2,h} \vh_i \|^2/2T}$ and $Z_h = \sum_{i=1}^{N} e^{- \| \mW_{1,h} \vz - \mW_{2,h} \vh_i \|^2/2T}$. In addition, the local minima of $\tilde{F}$ are attained when $\|\mW_{1,h}\vz\| = \rho$.	
\end{lemma}
\begin{proof}
	Recalling that 
	$$
	F =- \frac{T}{H} \sum_{h=1}^{H} \log \sum_{i=1}^{N} e^{- \frac{\| \mW_{1,h} \vz - \mW_{2,h} \vh_i \|^2}{2T}}. 
	$$ 
	We can compute the derivative of $F$ with respect to $\vz$ as
	\begin{equation*}
		\nabla_{\vz} F = \frac{1}{H} \sum_{h=1}^{H}\sum_{i=1}^N p_{i,h} \mW_{1,h}^T \left( \mW_{1,h} \vz - \mW_{2,h} \vh_i \right),
	\end{equation*}
	where $p_{i,h} = \frac{1}{Z_h}e^{- \frac{\| \mW_{1,h} \vz - \mW_{2,h} \vh_i \|^2}{2T}}$ and $Z_h = \sum_{i=1}^{N} e^{- \frac{\| \mW_{1,h} \vz - \mW_{2,h} \vh_i \|^2}{2T}}$.
	Since the attention heads are independent of each other, the proof for each head is similar to that of Lemma \ref{lemma:minima}.	
	We denote $\vr_{i,h} = \mW_{1,h}^T\left( \mW_{1,h}\vz - \mW_{2,h} \vh_i \right)$ and to compute the Hessian matrix, we first calculate
	\begin{equation*}
		\begin{aligned}
			\nabla_{\vz} p_{i,h} = - \frac{1}{T} p_{i,h} \vr_{i,h} + \frac{1}{T} p_{i,h} \sum_{j=1}^N p_{j,h} \vr_{j,h}. 
		\end{aligned}
	\end{equation*}
	Then the Hessian matrix of $F$ with respect to $\vz$ is
	\begin{equation*}
		\begin{aligned}
			\nabla^2_{\vz} F =  \frac{1}{H}\sum_{h=1}^H \Bigg[  \mW_{1,h}^T \mW_{1,h}  - \frac{1}{T} \mC_h \Bigg],
		\end{aligned}
	\end{equation*}
	where $\mC_h = \sum_{i=1}^N p_{i,h} \vr_{i,h} \vr_{i,h}^T -  \left(\nabla_{\vz} F_h\right) \left(\nabla_{\vz} F_h \right)^T$, $F_h^*$ is the Helmholtz free energy in the $h$-th subspace and $\nabla_{\vz} F_h^* = \sum_{i=1}^N p_{i,h} \vr_{i,h}$.
	As $\mW_{1,h}^T \mW_{1,h} \succeq 0$ and $-C_h \preceq 0$, we obtain that $F$ is neither convex nor concave and when $\| \vz \| \le \rho$, its local minima can only be attained at the boundary $\| \vz \| = \rho $ or at interior points where $\nabla_{\vz} F = 0$, that is, $\sum_{h=1}^H \sum_{i=1}^N p_{i,h} \left( \mW_{1,h}\vz - \mW_{2,h} \vh_i \right) = 0$.
	
	Similarly, we can obtain the Hessian matrix of $\tilde{F}$ as $\nabla^2_{\vz} \tilde{F} =  - \frac{1}{HT} \sum_{h=1}^H \left[ \sum_{i=1}^N p_{i,h} \vr_{i,h} \vr_{i,h}^T -  \left(\nabla_{\vz} \tilde{F}_h\right) \left(\nabla_{\vz} \tilde{F}_h \right)^T \right] \preceq 0$
	where $p_{i,h} = \frac{e^{\vz^T \mW_{1,h}^T\mW_{2,h} \vh_i / T}}{Z_h}$ and $Z_h$ is the normalizing constant. 
	Therefore, we can get that $\tilde{F}$ is concave and when $\| \mW_{1,h}\vz \| \le \rho$, its local minima can only be attained at the boundary $\| \mW_{1,h}\vz \| = \rho $.
\end{proof}

\begin{figure*}[t]
	\centering
	\includegraphics[scale=1.30]{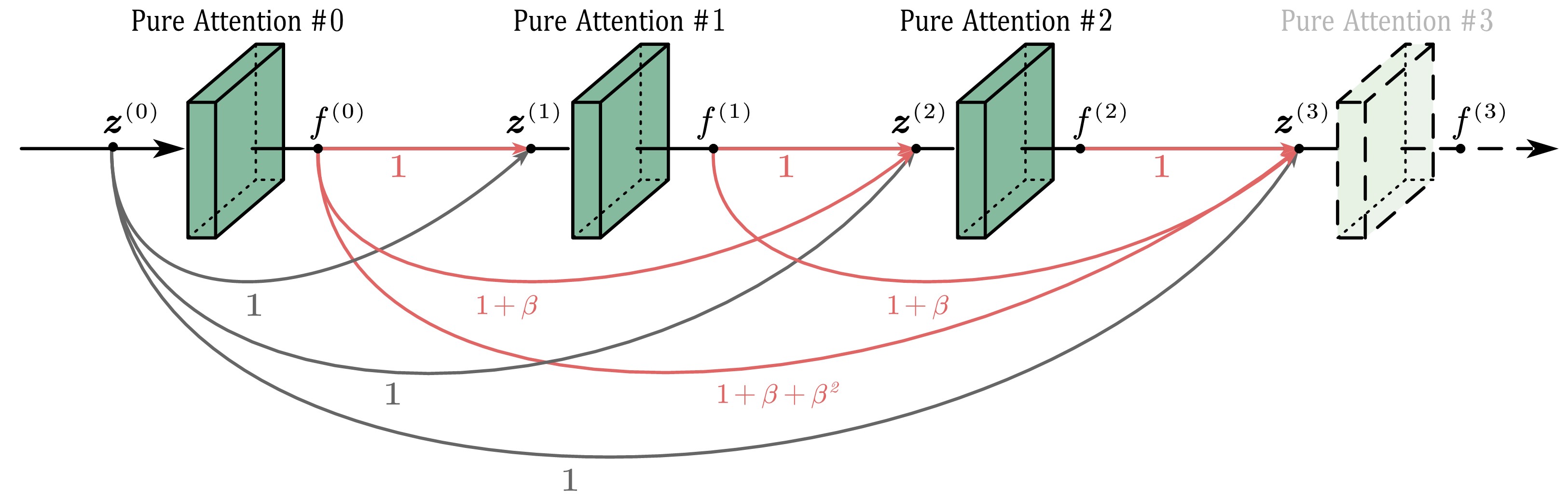}
	\caption{The illustration of the skip connections induced by ${\rm MomenMHA}$. We show the computation of $\vz^{(L)}$ in the case $L=3$ and $\eta = 1$. 
		The output $\vz^{(k)}$ of the $k$-th layer can be viewed as a weighted sum of the initial input $\vz^{(0)}$ (shown in gray) and the pure attention outputs $f^{(k)}$ from the previous $k-1$ layers (shown in pink). 
		The weight of each skip connection is indicated next to it.	
	}
	\label{fig:momen}
\end{figure*}

\subsection{More Discussions on ${\rm MomenMHA}$ and ${\rm NagMHA}$}\label{app:MomenMHA}

We use $f^{(k)}$ to denote the pure attention part (without skip connection), that is, 
\begin{equation*}
	f^{(k)} = \sum_{h=1}^{H} \sum_{i=1}^{N}  \frac{e^{(\vz^{(k)})^T \mW_{Q,h}^T \mW_{K,h} \vh_i /T }}{Z'_h} \mW_{O,h}\mW_{V,h} \vh_i.
\end{equation*}
We replace $ - \nabla_{\vz^{(k)}} F = f^{(k)} $ in Eq (\ref{eq:momenGD}) and Eq (\ref{eq:nagGD}) to obtain ${\rm MomenMHA}$ and ${\rm NagMHA}$ respectively.

As for ${\rm MomenMHA}$, in fact, we can expand the update of $\vz$ in Eq (\ref{eq:momenGD}) to obtain that
\begin{equation*}
\begin{aligned}
	\vz^{(L)} &= \vz^{(0)} + \sum_{k=0}^{L-1} \sum_{n=0}^{L-1-k} \eta \beta^n  \left( - \nabla_{\vz^{(k)}} F \right) \\
	&= \vz^{(0)} + \sum_{k=0}^{L-1} \left(\sum_{n=0}^{L-1-k} \eta \beta^n \right) f^{(k)}.
\end{aligned}
\end{equation*}
We can see that the output of the $L$-th layer can be viewed as a weighted sum of the initial input $\vz^{(0)}$ and the pure attention outputs $f^{(k)}$ from all previous $L-1$ layers with a weight following geometric series in $\beta^n$.
When $\beta = 0$, it degenerates to the case of standard residual connection.
This can be seen in Figure \ref{fig:momen}.
At first glance, this unrolled formulation bears a resemblance to DenseNet\cite{densenet}; however, it is crucial to highlight the fundamental differences. 
In DenseNet, feature maps from preceding layers are concatenated and directly fed into the next layer, allowing for the lossless utilization of prior information. 
In contrast, for ResNet\cite{resnet} (corresponding to the case where $\beta=0$), information is aggregated via direct summation with uniform weights. 
Consequently, as network depth increases, signals from earlier layers may become diluted or less significant.
However, by incorporating momentum, we alter this dynamic: the weights of earlier layers are progressively amplified by a factor of $(1+\beta^n)$, with $n$ being the layer interval.
In practice, we parameterize $\beta$ using a lightweight two-layer scalar gate
$\beta(\vz)=\sigma\left(\operatorname{SiLU}\left(\vz^T\mA \right)\mB \right)$,
where $\mA \in\mathbb{R}^{d\times m_{\beta}d}$, $\mB_h\in\mathbb{R}^{m_{\beta}d\times1}$, and $m_{\beta}$ is a factor.
Meanwhile, we set $\eta$ as learnable parameters initialized to 1.0 for each layer.

\begin{figure*}[t]
	\centering
	\includegraphics[scale=1.30]{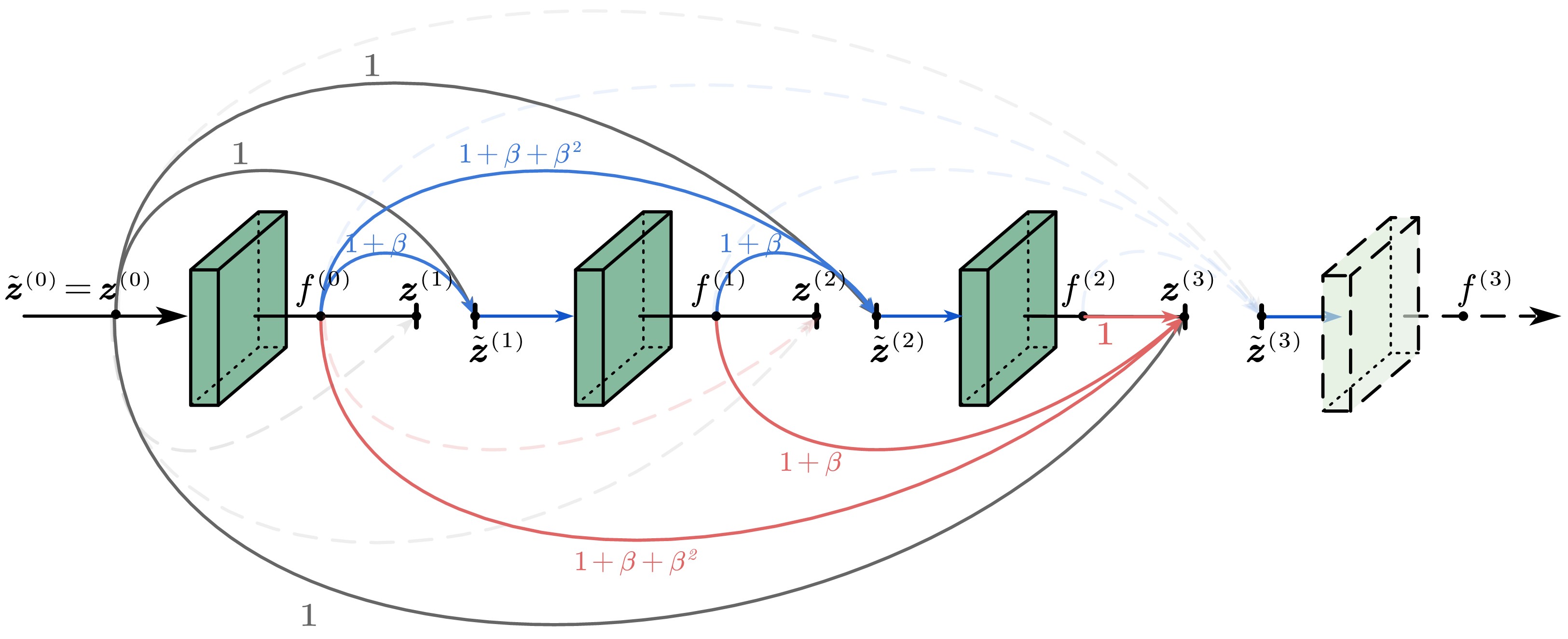}
	\caption{The illustration of the skip connections induced by ${\rm NagMHA}$. We show the computation of $\vz^{(L)}$ in the case $L=3$ and $\eta = 1$. 
		Solid lines denote the skip connections that actually contribute to $\vz^{(3)}$, and the arrow directions indicate the flow of information. 
		By introducing lookahead output $\tilde{\vz}^{(k)}$, additional nodes are incorporated into the information flow, resulting in denser extra skip connections compared to  ${\rm MomenMHA}$.	
	}
	\label{fig:nag-simple}
\end{figure*}

As for ${\rm NagMHA}$, we can also expand $\tilde{\vz}$ and $\vz$ to obtain 
\begin{align*}
	\tilde{\vz}^{(L)} &= \vz^{(0)} - \sum_{k=0}^{L-1}\left(\eta \sum_{n=0}^{L-1-k}\beta^n + \beta^{L-k} \right) \nabla_{\tilde{\vz}^{(k)}} F \\
	&= \vz^{(0)} + \sum_{k=0}^{L-1}\left(\eta \sum_{n=0}^{L-1-k}\beta^n + \beta^{L-k} \right) \tilde{f}^{(k)}, \\
	\vz^{(L)} &= \vz^{(0)} + \sum_{k=0}^{L-1} \sum_{n=0}^{L-1-k} \eta \beta^n  \left( - \nabla_{\tilde{\vz}^{(k)}} F \right) \\
	&= \vz^{(0)} + \sum_{k=0}^{L-1} \left(\sum_{n=0}^{L-1-k} \eta \beta^n \right) \tilde{f}^{(k)},
\end{align*}
respectively. When we focus only on the final output $\vz^{(L)}$, the skip connections that actually contribute to $\vz^{(L)}$ is shown in Figure \ref{fig:nag-simple}.
By incorporating lookahead representations $\tilde{\vz}^{(k)}$ as additional nodes within the information flow, this mechanism introduces denser skip connections.
In practice, the remaining design details and discussions are similar to those in ${\rm MomenMHA}$, and are thus omitted for brevity.

\subsection{Detailed Design of ${\rm DiagMHA2nd}$}\label{app:diag2nd}
In addition to first-order momentum-based methods, another simple and straightforward idea for employing a more efficient algorithm is to use Newton’s method, which can be formulated as
\begin{equation*}
	\vz^{(k+1)} = \vz^{(k)} - \eta  \left[ \nabla_{\vz^{(k)}}^2 F \right]^{-1}  \nabla_{\vz^{(k)}} F,
\end{equation*}
where $\nabla_{\vz^{(k)}}^2 F$ is the Hessian matrix at $\vz^{(k)}$.
The above update can be viewed as preconditioning the gradient with the second-order information from the Hessian matrix.

To identify the induced second-order attention structure before introducing learnable projections, we first consider an unparameterized subspace.
Let $\vz^{(h)}\in\mathbb{R}^{d_h}$ denote the token to be updated in the $h$-th subspace and let $\vu_i^{(h)}\in\mathbb{R}^{d_h}$ denote the corresponding component in the $h$-th subspace derived from the original interacting token $\vh_i$.
We denote the Helmholtz free energy in the $h$-th subspace as $F_h= -T\log\sum_{i=1}^{N} \exp\left( (\vz^{(h)})^T\mathbf{u}_i^{(h)} / T \right)$ and then the global energy is $F = \frac{1}{H}\sum_{i=1}^H F_h$.

Let $\mR_h\in\mathbb{R}^{d_h\times d}$ denote the mapping from the original representation to the $h$-th subspace, i.e., $\vz^{(h)}=\mR_h\vz$. 
Then the Hessian of the global energy can be written as
\begin{equation*}
\nabla_{\vz}^{2}F = \frac{1}{H} \sum_{h=1}^{H} \mR_h^T \nabla_{\vz^{(h)}}^{2}F_h \mR_h.
\end{equation*}
It can be noticed that each term $\mR_h^T\nabla_{\vz^{(h)}}^{2}F_h\mR_h\in\mathbb{R}^{d\times d}$ has rank at most $d_h$ as $\mR_h\in\mathbb{R}^{d_h\times d}$ with $d_h<d$.
Consequently, the global Hessian is obtained by aggregating multiple low-rank components.
Their aggregation is not guaranteed to be invertible and can be poorly conditioned, while directly inverting the global Hessian also couples the second-order structures from different subspaces.
Considering that multi-head attention itself decomposes the representation into separate head-wise subspaces, a more natural and tractable choice is to perform the second-order preconditioning independently within each subspace.

Therefore, instead of applying Newton’s method directly to $F$, we apply it independently to each subspace $F_h$, that is,
\begin{equation*}
	\vz^{(k+1)} = \vz^{(k)} - \frac{\eta}{H}\sum_{h=1}^H \left[ \nabla_{\vz^{(k)}}^2 F_h \right]^{-1}  \nabla_{\vz^{(k)}} F_h.
\end{equation*}

Denoting $\bar{\mathbf{u}}^{(h)} = \sum_i p_{i,h}\mathbf{u}_i^{(h)}$ where the normalized weight is $p_{i,h} = \operatorname{softmax}_i \big((\vz^{(h)})^T\mathbf{u}_i^{(h)}/T\big)$,
we have $\nabla_{\vz^{(h)}}F_h=-\bar{\mathbf{u}}^{(h)}$ and the Hessian matrix is
\begin{equation*}
    \nabla_{\vz^{(h)}}^2F_h = -\frac{1}{T} \sum_{i=1}^{N} p_{i,h}
	\left( \mathbf{u}_i^{(h)}-\bar{\mathbf{u}}^{(h)} \right) 
    \left( \mathbf{u}_i^{(h)}-\bar{\mathbf{u}}^{(h)} \right)^T.
\end{equation*}
For notation simplify, we denote the attention-weighted covariance part as $\mC_h = - T \nabla_{\vz^{(h)}}^2F_h$.
It can be seen that this part naturally characterizes the local second-order geometry.
In particular, a larger curvature along a certain direction indicates that the energy changes more rapidly along this direction, for which a Newton-style preconditioner should produce a relatively more conservative update.
The key to incorporating such second-order information is therefore to use the curvature encoded by $\mC_h$ to precondition the first-order update.
Noticing that the global sign factors can be absorbed into the subsequent learnable projections, we focus on the inverse-curvature structure and use $\mC_h^{-1}$ as the basic form of the preconditioner.


To further improve invertibility and numerical stability, a common strategy in Newton-type optimization is to add an identity matrix to the curvature matrix \cite{bonnans2006numerical}. 
Accordingly, we first consider the regularized preconditioner 
\begin{equation*}
\mP_h = \left( \lambda_h\mI+\mC_h \right)^{-1}.
\end{equation*}
The coefficient $\lambda_h$ controls the damping strength and improves the conditioning of the curvature matrix, thereby promoting its invertibility.
By factoring out $\lambda_h$, we have $\left(\lambda_h\mI+\mC_h\right)^{-1} = \frac{1}{\lambda_h} \left(
\mI+\frac{1}{\lambda_h}\mC_h \right)^{-1}.$
The overall scalar factor $1/\lambda_h$ only uniformly rescales the update and can be absorbed into the subsequent learnable projection.
We therefore focus on the relative curvature-dependent scaling and reparameterize $1/\lambda_h$ as $\alpha_h$, yielding
\begin{equation*}
\mP_h = \left( \mI+\alpha_h\mC_h \right)^{-1},
\end{equation*}
where $\alpha_h>0$ controls the relative strength of the curvature-dependent correction.
Consequently, directions with larger estimated curvature are assigned smaller effective update magnitudes.

Noticing that directly constructing $\mC_h\in\mathbb{R}^{d_h\times d_h}$ and computing its inverse for every query introduce considerable computational and memory overhead, we further simplify the second-order structure.
For all $N$ query tokens, explicitly constructing the full covariance matrices requires
$O(N^2 H d_h^2)=O(N^2 d d_h)$ computation and
$O(NH d_h^2)=O(N d d_h)$ additional storage, while performing matrix inversion for every query further incurs
$O(NH d_h^3)$ computation.
In comparison, standard multi-head attention has a computational complexity of
$O(N^2d+Nd^2)$.
Therefore, the full covariance formulation introduces a substantial additional cost, especially for long sequences, making it difficult to use in practice.
We thus approximate the matrix using only its diagonal entries \cite{saad2003iterative}, that is,
\begin{equation*} 
    \mC_h \approx \operatorname{Diag}(\vs_h),
\end{equation*}
\begin{equation*}
	\vs_h = \sum_{i=1}^{N} p_{i,h} \left( \vu_i^{(h)} \odot \vu_i^{(h)} \right) - \bar{\vu}^{(h)} \odot \bar{\vu}^{(h)}.
\end{equation*}
Here $\vs_h$ is exactly the coordinate-wise variance.
Compared with the full covariance matrix, this diagonal approximation retains the curvature magnitude along each feature dimension while discarding the cross-dimensional correlations.
More importantly, its inverse can be evaluated element-wise without explicitly constructing or inverting a matrix.

We next instantiate the above second-order structure into the attention modification.
As in standard attention, the query and key projections are used to determine the interaction weights,
\begin{equation*}
	p_{i,h} = \operatorname{softmax}_i \left( \frac{ \vz^T \mW_{Q,h}^T \mW_{K,h} \vh_i}{T} \right).
\end{equation*}
For the space on which the update is actually performed, we use the value projection
$\vv_{i,h}=\mW_{V,h}\vh_i$.
Accordingly, the standard first-order attention update in the $h$-th head is
$\bar{\vv}_h=\sum_i p_{i,h}\vv_{i,h}$ while its diagonal second-order statistic becomes
\begin{equation*}
	\vs_h = \sum_{i=1}^{N} p_{i,h} \left(
	\vv_{i,h}\odot\vv_{i,h} \right) - \bar{\vv}_h\odot\bar{\vv}_h .
\end{equation*}
In this formulation, $\mW_Q$ and $\mW_K$ determine which tokens contribute to the update through the attention distribution, whereas $\mW_V$ defines the representation space in which both the first-order update direction and its second-order curvature are measured.

\begin{figure*}[t]
    \centering
    \begin{minipage}[t]{0.49\textwidth}
        \centering
        \includegraphics[width=\linewidth]
        {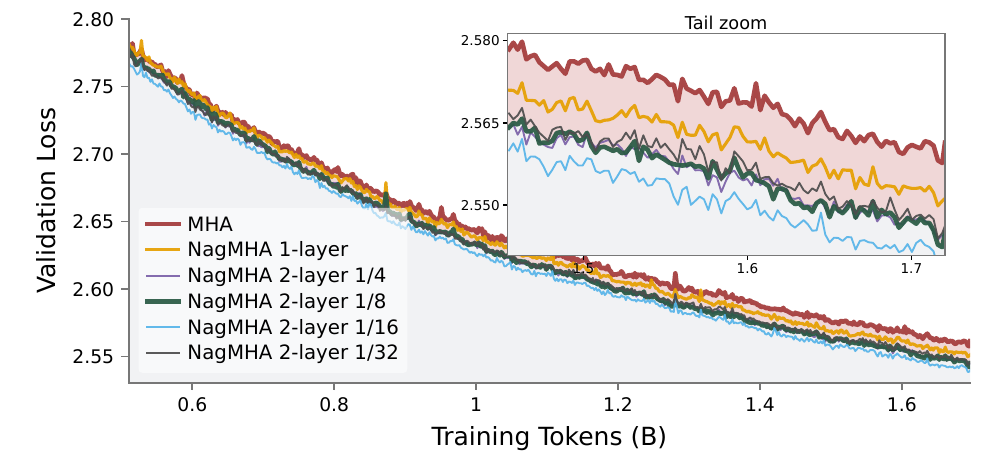}

        \vspace{-2mm}
        {\small (a) NagMHA - Training Tokens}
    \end{minipage}
    \hfill
    \begin{minipage}[t]{0.49\textwidth}
        \centering
        \includegraphics[width=\linewidth]
        {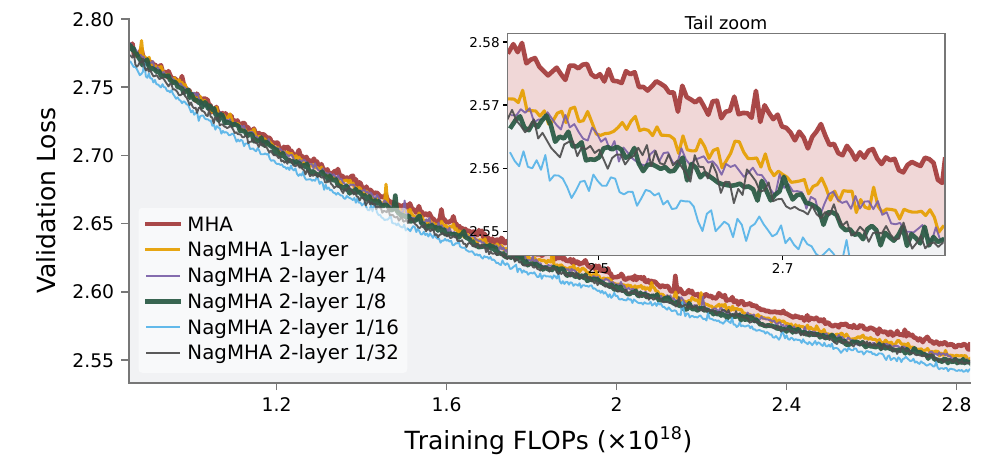}

        \vspace{-2mm}
        {\small (b) NagMHA - Training FLOPs}
    \end{minipage}

    \vspace{1mm}

    \begin{minipage}[t]{0.49\textwidth}
        \centering
        \includegraphics[width=\linewidth]
        {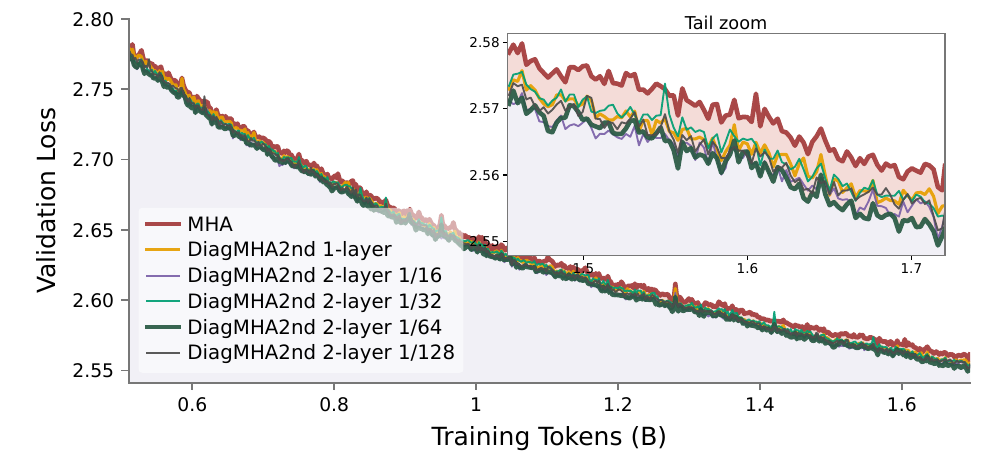}

        \vspace{-2mm}
        {\small (c) DiagMHA2nd - Training Tokens}
    \end{minipage}
    \hfill
    \begin{minipage}[t]{0.49\textwidth}
        \centering
        \includegraphics[width=\linewidth]
        {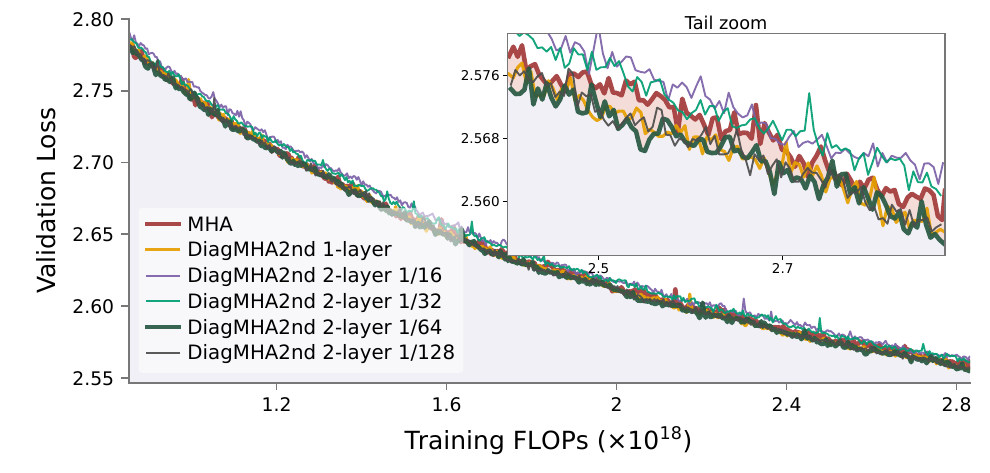}

        \vspace{-2mm}
        {\small (d) DiagMHA2nd - Training FLOPs}
    \end{minipage}

    \vspace{-1mm}
    \caption{
        \textbf{Training dynamics under different gating configurations of different gating configurations for 300M NagMHA and DiagMHA2nd.}
        (a,b) Comparison of different $\beta$-gate configurations in NagMHA.
        (c,d) Corresponding comparison of $\alpha$-gate configurations in DiagMHA2nd.
        The fractions in the legends denote the hidden dimension factors $m_{\beta}$ and $m_{\alpha}$ for the two-layer gates, where $d_{\beta}=m_{\beta}d$ and $d_{\alpha}=m_{\alpha}d$, respectively.
        The single-layer gate and standard MHA are included for comparison.
        The inset plots provide enlarged views of the final training stages.
        For NagMHA, different gating configurations generally outperform standard MHA under both matched training tokens and matched FLOPs, with the two-layer gate at $m_{\beta}=1/16$ achieving the lowest validation loss.
        For DiagMHA2nd, the two-layer gate at $m_{\alpha}=1/64$ exhibits favorable performance.
        Furthermore, increasing the gate hidden dimension beyond these settings does not consistently yield additional improvements under matched training FLOPs.
    }
    \label{fig:gate_ablation_dynamics}
    \vspace{-2mm}
\end{figure*}

In addition, since the diagonal preconditioner is primarily concerned with the relative curvature across different dimensions while the absolute magnitude of $\vs_h$ can vary with the scale of the value representations, we therefore normalize $\vs_h$ by its mean within each head, that is, $\widetilde{\vs}_h =  {\operatorname{sg}(\vs_h)}/({ \mu_h^{(s)}+\epsilon}$) where $\mu_h^{(s)} = \operatorname{mean}
\big( \operatorname{sg}(\vs_h) \big)$ is the mean for each head and $\operatorname{sg}(\cdot)$ denotes the stop-gradient operator.
This normalization removes the common scale of the estimated curvature while preserving the relative variation among different feature dimensions.
In addition, the stop-gradient operation is used to treat the second-order statistic as a numerical preconditioning signal during the forward pass, avoiding an additional backward optimization path through its estimation.

Furthermore, we make the strength of the second-order correction depend on the current token and parameterize $\alpha_h$ using a lightweight two-layer scalar gate,
$\alpha_h(\vz)=\sigma\left(\operatorname{SiLU}\left(\vz^T\mA_h\right)\mB_h\right)$,
where $\mA_h\in\mathbb{R}^{d\times m_{\alpha}d}$, $\mB_h\in\mathbb{R}^{m_{\alpha}d\times1}$, and $m_{\alpha}$ is a factor with $m_{\alpha}d \ll d$.
It can be noticed that $\alpha_h$ is a scalar shared by all feature dimensions within the same head.
Therefore, the coordinate-wise geometry of the preconditioner is still determined by the analytically estimated second-order statistic $\widetilde{\vs}_h$, while the learnable gate only controls the overall strength of the second-order correction.

Combining the above components, we obtain our final attention modification based on second-order information as
\begin{equation*}
	\begin{aligned}
		{\rm DiagMHA2nd}(\vz) &= \vz + \sum_{h=1}^{H} \mW_{O,h} \mD_h \bar{\vv}_h,
		\\
		\mD_h &=\left[\mI+\alpha_h(\vz)\operatorname{Diag}\left(\widetilde{\vs}_h\right)\right]^{-1}.
	\end{aligned}
\end{equation*}
Noticing that $\mD_h$ is diagonal, the preconditioning can be implemented element-wise.
Therefore, ${\rm DiagMHA2nd}$ can be equivalently written as
\begin{equation*}
{\rm DiagMHA2nd}(\vz) = \vz + \sum_{h=1}^{H} \mW_{O,h}
\frac{ \bar{\vv}_h}{1+\alpha_h(\vz)\widetilde{\vs}_h},
\end{equation*}
where the division is performed element-wise.


Compared with standard MHA, DiagMHA2nd mainly introduces the computation of the diagonal second-order statistic.
Specifically, computing
$\sum_i p_{i,h}(\vv_{i,h}\odot\vv_{i,h})$
requires one additional attention-weighted aggregation.
Since this operation has the same form as the standard value aggregation
$\sum_i p_{i,h}\vv_{i,h}$,
its computational cost over all heads and query tokens is $O(N^2d)$.
The subsequent computation of the coordinate-wise variance, mean normalization, and diagonal preconditioning only involves element-wise operations, resulting in an additional cost of $O(Nd)$.
The two-layer scalar gate introduces $O(NHdm)$ computation, where $m\ll d$ is the intermediate dimension, and its cost is therefore relatively small and at most of order $O(Nd^2)$ under our setting.
By using the diagonal approximation, DiagMHA2nd avoids the $O(N^2dd_h)$ computation required to explicitly construct the full covariance matrices, as well as the additional $O(NHd_h^3)$ cost of performing matrix inversions for all queries.
The remaining second-order computation is reduced to an additional $O(N^2d)$ attention-weighted aggregation together with lightweight element-wise operations.
Therefore, including the original attention and linear projections, the overall computational complexity of DiagMHA2nd is $O(N^2d+Nd^2)$ which has the same asymptotic order as standard multi-head attention.

\subsection{More Details of Experiments}\label{app:Ex}
To explore the potential of the proposed attention modifications, we conduct experiments using a GPT-like architecture\cite{GPT2}. 
Specifically, we replace the original standard Softmax attention with the ${\rm MomenMHA}$ and ${\rm NagMHA}$ introduced in Section \ref{sec:atten1st}, as well as the ${\rm DiagMHA2nd}$ described in Section \ref{sec:atten2nd} while keeping the remaining architecture unchanged.
For the FFN blocks, we use GELU \cite{GELU} as the activation function, and the hidden layer dimension is 4 times the input dimension.
Relative positional bias is enabled for all models. All configurations use a head dimension of $d_h=64$ and a maximum context length of 768 tokens.

We conduct pre-training on the MiniPile dataset \cite{minipile}, which is a compact subset version of the original Pile dataset \cite{Pile}.
The original MiniPile splits contain 1,000,000 training documents. 
We tokenize each document using the GPT-2 BPE tokenizer provided by Hugging Face~\cite{huggingface} and and manually append an end-of-text token to each document.
The resulting token streams are concatenated and divided into non-overlapping windows of 768 tokens.
Incomplete tails are discarded, and no input padding is applied.
This preprocessing produces 2.21M training windows, corresponding to around 1.7B input tokens per traversal of the packed training split.
Training uses the standard causal next-token prediction objective.
The effective batch size is 16 sequences corresponding to 12,288 input tokens for a regular optimization step.

The model sizes range from 30M, 50M, 70M, 160M and 300M parameters.
These models have layers and attention heads in the ranges $\{ 6,8,8,12,16\}$ and $\{4,6,8,12,16\}$ respectively, with each head of dimensionality $d_h = 64$.
Since the gating networks in attention variants introduce only a small number of additional parameters, the parameter counts of different attention variants remain comparable at each model scale, with a maximum difference of approximately 2\%.
For $\rm {MomenMHA}$ and $\rm {NagMHA}$, we set the hidden dimension for the 2-layer mlp gate as $m_{\beta} = 1/8$ for 300M families while $m_{\beta} = 1/4$ for remaining families.
To maintain comparable parameter counts, we set the hidden dimension ratios of the $\alpha$-gate in DiagMHA2nd to $m_{\alpha}=1/8, 1/12, 1/16, 1/24,$ and $1/64$ for the five model sizes respectively.
The 30M, 50M, and 70M families are trained for one epoch, whereas the 160M and 300M families are trained for three epochs, corresponding to approximately 5.1B input tokens.
For all models except the 160M one, we set the batch size to 32; for the 160M model, the batch size is set to 16.
We use AdamW \cite{AdamW} as the optimizer with  $\beta_1 = 0.9$, $\beta_2 = 0.999$.
The learning rate is selected according to model scale.
We performed a learning rate sweep for standard MHA and identified the optimal learning rates as $3\times10^{-3}$, $1\times10^{-3}$, $1\times10^{-3}$, $5\times10^{-4}$, and $3\times10^{-4}$ for the 30M, 50M, 70M, 160M, and 300M models, respectively. We also conducted learning rate sweeps for the other three attention variants and found that the optimal learning rates identified for MHA were equally suitable for these variants.
All experiments are implemented on four 40G A100 GPUs.

\bibliographystyle{IEEEtran}
\bibliography{sample-base}

\end{document}